\documentclass[10pt,twocolumn,letterpaper]{article}

\usepackage{cvpr}              

\usepackage{graphicx}
\usepackage{amsmath}
\usepackage{amssymb}
\usepackage{booktabs}
 \usepackage{multirow}
\usepackage{enumitem}
\usepackage[title]{appendix}

\usepackage[table,xcdraw]{xcolor}

\usepackage[pagebackref,breaklinks,colorlinks]{hyperref}

\usepackage[capitalize]{cleveref}
\crefname{section}{Sec.}{Secs.}
\Crefname{section}{Section}{Sections}
\Crefname{table}{Table}{Tables}
\crefname{table}{Tab.}{Tabs.}

\makeatletter
\def\thanks#1{\protected@xdef\@thanks{\@thanks
        \protect\footnotetext{#1}}}
\makeatother

\begin{document}

\title{TCP:Textual-based Class-aware Prompt tuning for Visual-Language Model}

\author{\small Hantao Yao$^1$,Rui Zhang$^2$, Changsheng Xu$^{1,3}$\\
$^1$ \small State Key Laboratory of Multimodal Artificial Intelligence Systems, Institute of Automation, CAS\\
$^2$ \small State Key Lab of Processors, Institute of Computing Technology, CAS; $^3$ \small University of Chinese Academy of Sciences(CAS),\\
{\tt\small hantao.yao@nlpr.ia.ac.cn}
}

\maketitle

\begin{abstract}
Prompt tuning represents a valuable technique for adapting pre-trained visual-language models (VLM) to various downstream tasks.
Recent advancements in CoOp-based methods propose a set of learnable domain-shared or image-conditional textual tokens to facilitate the generation of task-specific textual classifiers. 
However, those textual tokens have a limited generalization ability regarding unseen domains, as they cannot dynamically adjust to the distribution of testing classes.
To tackle this issue, we present a novel Textual-based Class-aware Prompt tuning(TCP) that explicitly incorporates prior knowledge about classes to enhance their discriminability.
The critical concept of TCP involves leveraging Textual Knowledge Embedding (TKE) to map the high generalizability of class-level textual knowledge into class-aware textual tokens.
By seamlessly integrating these class-aware prompts into the Text Encoder, a dynamic class-aware classifier is generated to enhance discriminability for unseen domains.
During inference, TKE dynamically generates class-aware prompts related to the unseen classes.
Comprehensive evaluations demonstrate that TKE serves as a plug-and-play module effortlessly combinable with existing methods. 
Furthermore, TCP consistently achieves superior performance while demanding less training time.Code:\url{https://github.com/htyao89/Textual-based_Class-aware_prompt_tuning/}
\end{abstract}


\section{Introduction}
\label{sec:intro}
Large-scale image-text pairs have the ability to train a Visual-language model (VLM) with a powerful generalization capacity for various downstream tasks~\cite{RadfordKHRGASAM21,abs-2204-14198}. However, training these models from scratch requires a huge dataset with labeled images, making it difficult to apply them directly to downstream tasks with fewer images.
To address this issue, three common techniques have been proposed: fine-tuning~\cite{DBLP:conf/cvpr/OquabBLS14}, prompt tuning~\cite{ZhouYLL22}, adapter~\cite{DBLP:conf/iclr/HeZMBN22} and LoRA~\cite{HuSWALWWC22}. 
Among them, prompt tuning is a simple and efficient framework that transfers the essential general knowledge of VLM to the downstream tasks.

\begin{figure}
  \centering
   \includegraphics[width=0.95\linewidth]{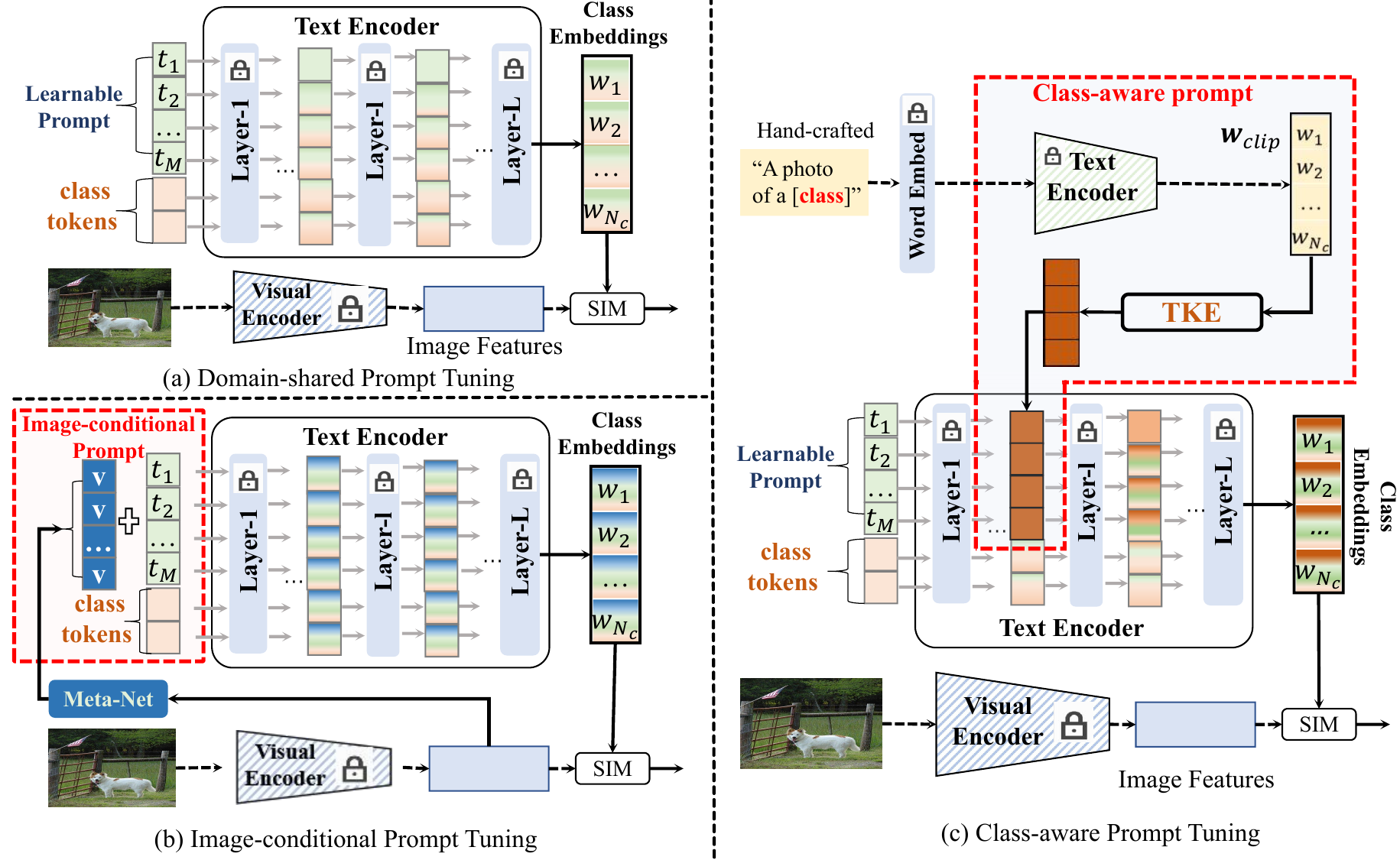}
   \caption{Comparison with existing frameworks. (a) Domain-shared prompt tuning applies the same learnable prompt between the training and testing domains. (b) Image-conditional prompt tuning combines the image embedding with the learnable prompt; (c) Class-aware Prompt tuning injects the class-level textual embedding into the Text Encoder with the class-aware prompt.}
   \label{fig:motivaiton}
\vspace{-1.50em}
\end{figure} 

Prompt tuning\footnote{For the visual-language model, two types of prompt-tuning strategies exist: visual prompt tuning and textual prompt tuning. In this work, we focus on the textual prompt tuning and do not consider the visual prompt tuning.} is a technique that combines learnable textual tokens with class tokens to generate a discriminative textual classifier, known as Context Optimization (CoOp)~\cite{ZhouYLL22}.
Recently, various CoOp-based methods~\cite{ZhouYLL22,YaoZX23,0002YSLR023,BlackBox,KAPT23,DAPT23,RPO23,MAPLE23,DBLP:journals/corr/abs-2303-15234,PromptReg23} infer the domain-shared prompt tokens between training and testing domains(Figure~\ref{fig:motivaiton}(a)).
Nevertheless, as the domain-shared prompt tokens are derived from labeled training images, their performance is suboptimal when confronted with unseen test classes.
To enhance the generalization capacity of learnable prompt tokens, image-conditional prompts have been proposed in~\cite{ZhouYL022,abs-2210-07225} by fusing image features and learnable tokens (Figure~\ref{fig:motivaiton}(b)).
Notably, image-conditional textual tokens encapsulate specific knowledge of each image, particularly for testing images,thereby rendering it more apt for generalization to unseen testing images. 
However, image-conditional prompt tokens with image-specific knowledge have less ability in elevating the distribution of class embeddings.
In summary, classifiers generated by domain-shared and image-conditional textual tokens exhibit suboptimal performance for the unseen classes, primarily due to their inability to explicitly model the class distribution. 
Consequently, it is imperative to establish a dynamic relationship between the learnable prompt and the textual knowledge of each class to augment its discriminative prowess.

The frozen CLIP, coupled with a hand-crafted prompt, exhibits robust generalization capabilities to novel classes, rendering it a valuable source of prior textual knowledge for each class.
By associating the class-level textual knowledge with the learnable prompts, a class-aware prompt can be formulated for hencing the discriminative capacity of the textual classifier.
To achieve this, we employ an embedding module to project the class-aware textual knowledge into the class-aware prompt tokens, as shown in Figure~\ref{fig:motivaiton}(c).
The resultant class-aware prompt incorporates prior textual knowledge specific to each class, endowing the generated textual classifier with heightened discriminative prowess.
Furthermore, the class-aware prompt facilitates the generation of classifiers for both seen and unseen classes by leveraging textual knowledge from both categories. 
To sum up, the trained embedding module can generate a class-aware prompt for each class based on its description (`class name'), thereby enhancing generalization and discriminative capabilities in class-level textual embeddings.

Therefore, we propose a novel Textual-based Class-aware Prompt tuning (TCP) based on the framework of CoOp, shown in Figure~\ref{fig:TCP}. 
In addition to the domain-share textual tokens introduced in CoOp, TCP contributes a novel Textual Knowledge Embedding (TKE) to map the class-level textual knowledge into class-aware prompt tokens. 
Moreover, a class-aware textual classifier is generated by inserting the class-aware prompt tokens into the middle layer of the Text Encoder. 
We use standard contrastive loss and knowledge-guided consistency~\cite{YaoZX23} to optimize the TKE and the learnable prompt tokens. 
During inference, TCP generates a class-aware classifier for unseen classes by feeding the domain-shared prompt tokens and the class-aware prompt tokens generated by the TKE into the frozen Text Encoder. 

Overall, the proposed TCP explicitly steers prompts to learn a class-aware knowledge that maximizes the generalization and discriminative of the downstream tasks.
Evaluation over 11 image classification datasets on base-to-new generalization, cross-dataset generalization, and few-shot learning verify shows that TCP is an efficient method that \emph{obtain a higher performance with less training time}.
In summary, the proposed Textual-based Class-aware Prompt tuning(TCP) has the following main contributions:
\begin{enumerate}
\item An effectively Textual-based Class-aware Prompt tuning is proposed by injecting the textual class-aware prompts generated by Textual Knowledge Embedding(TKE) into the Text Encoder.
\item We demonstrate that explicitly incorporating the prior knowledge of each class into the learnable prompt tokens can enhance the discriminative of the class distribution.
\item Textual Knowledge Embedding(TKE) is a plug-and-play module that can quickly insert existing prompt tuning methods to improve their performance further.
\end{enumerate}

\section{Related Works}
\label{sec:related}
\subsection{Vision-Language Models}
Recently, researchers have shown that the Visual-Languge Models(VLM)~\cite{RadfordKHRGASAM21,abs-2204-14198} consisting of visual and textual modalities trained on the large-scale of image-text pairs has a powerful generalization and discriminativeion.
To further boost the descriptive ability of VLM, the VLM models are boosted from the following aspect: 1) using a stronger text encoder or visual encoder~\cite{VaswaniSPUJGKP17,zhai2022scaling,li2023blip}; 2) deeply fusing the visual and text knowledges~\cite{li2022blip,singh2022flava}; 3) using more images~\cite{RadfordKHRGASAM21,JiaYXCPPLSLD21,schuhmann2021laion,schuhmann2022laion}.
To boost the diversity of text description, Masked Language Modeling (MLM) ~\cite{KimSK21}~\cite{LuBPL19} randomly erases the words in the text description used for representation learning.
Unlike MLM, Masked autoencoder-based methods~\cite{HeCXLDG22} are proposed to boost the descriptive ability by randomly masking the image patches.
Among existing VLM models, CLIP is a representative and straightforward framework for inferring the independent visual and text encoder using the contrastive loss based on 400 million image-text association pairs.
As CLIP has a good generalization, most existing CoOp-based methods are proposed based on CLIP for adapting the pre-trained VLM into the downstream task.
Similar to existing methods, we conduct the prompt-tuning strategy on the TextEncoder of CLIP to obtain a task-specific textual embedding for prediction.

\subsection{Prompt Tuning}
To adapt the pretrained VLM to the downstream tasks, the prompt tuning~\cite{abs-2210-09263,BlackBox,RPO23,abs-2210-07225,RadfordKHRGASAM21,DBLP:journals/corr/abs-2303-06571} always applies task-related textual tokens to infer the task-specific textual knowledge.
In CLIP\cite{RadfordKHRGASAM21}, the hand-crafted template ``a photo of a [CLASS]" is used to embed the textual embedding for zero-shot prediction.
However, the hand-crafted prompts have a poor ability to describe the downstream task.
Textual prompt tuning is applied to boost the textual embedding by inferring a set of learnable textual tokens combined with the class tokens.
For example, Context Optimization(CoOp)~\cite{ZhouYLL22} replaces the hand-crafted prompts with the learnable soft prompts.
To improve the generalization of the learnable textual prompt in CoOp, Conditional Context Optimization(CoCoOp)~\cite{ZhouYL022} and VPT~\cite{abs-2210-07225} generates an image-conditional prompt fusing the image feature and the learnable textual prompt.
Moreover, Knowlege-Guided Context Optimization(KgCoOp)~\cite{YaoZX23}, ProGrad~\cite{abs-2205-14865}, and Prompt Regularization(ProReg)~\cite{PromptReg23} constrain the proposed learnable prompts contain the essential general knowledge.
Unlike the above methods, which consider textual prompts, Ensembling Context Optimization(ECO)~\cite{ECO23} employs prompt ensembling to combine multiple prompts.
To obtain high-quality task-related tokens, ProDA~\cite{proda} considers the prompt's prior distribution learning, and Distribution-Aware Prompt Tuning (DAPT)~\cite{DAPT23} optimizes the learnable prompt by maximizing inter-dispersion.
Besides the textual knowledge from ``class-name", Knowlege-Aware Prompt Tuning(KAPT)~\cite{KAPT23} employs the external knowledge to generate the discriminative knowledge-aware prompt for the novel categories.
PLOT~\cite{0002YSLR023} applies optimal transport to match the vision and text modalities for generating the discriminative and visual-aligned local textual prompt tokens.
Besides the textual prompt tuning, Multi-modal Prompt Learning (MaPLe)~\cite{MAPLE23} and PromptSRC~\cite{DBLP:journals/corr/abs-2307-06948} conduct the visual-textual prompt tuning by jointly conducting the prompt tuning on the visual and text encoders.
Multitask Vision-Language Prompt Tuning(MVLPT)~\cite{DBLP:journals/corr/abs-2211-11720}incorporates cross-task knowledge into prompt tuning for
vision-language models.
DenseCLIP~\cite{RaoZ0TZH0L22} uses the context-aware prompt strategy to generate dense prediction tasks, and CLIP-Adapter~\cite{abs-2110-04544} applies an adapter to adjust the visual or text embeddings.

Existing methods commonly infer two types of prompt tokens: domain-share and image-conditional. 
However, the textual classifiers generated with these tokens tend to perform poorly on unseen classes. 
To mitigate this limitation, we propose a novel Textual-based Class-aware Prompt tuning (TCP) that employs a dynamic class-aware token to enhance the generalization and discriminative capabilities of the learnable textual prompt.
Moreover, we introduce a Textual Knowledge Embedding to project class-level textual knowledge into the class-aware prompts, which are then inserted into the Text Encoder to generate a discriminative class-aware classifier. 
The evaluation results demonstrate that integrating class-level prior knowledge into the prompt tokens significantly enhances the discriminative ability of the prompt tuning process.

\section{Methodolgy}
As the Textual-based Class-aware Prompt tuning(TCP) is proposed based on Context Optimization (CoOp), we first briefly review CoOp and then introduce the proposed TCP.

\subsection{ Preliminaries}
Existing CoOp-based methods are proposed based on the powerful Contrastive Language-Image Pre-training(CLIP).
Given an image along with its corresponding textual description, CLIP uses visual and text encoders to extract the visual and text embeddings.
After that, the constrastive loss between the visual and textual embeddings is calculated to align those two embeddings.
To effectively adapt CLIP for the downstream task, 
CLIP applies the hand-crafted template ``a photo of a \{\}" to extract the general class-level textual embedding, defined as $\mathbf{W}^{clip}=\{\mathbf{w}^{clip}_i\}_{i=1}^{{N}_c}$, where $\mathbf{w}^{clip}_i$ is the textual embedding of $i$-th class, and $N_c$ is the number of classes.
Given the `class-name' of $i$-th class, Word Embedded $e(\cdot)$ firstly embeds the hand-crafted description into a vectorized textual tokens: $\mathbf{t}^{clip}_i=e$(``a photo of a \{class-name\}''). 
After that, Text Encoder $\theta$ maps the vectorized textual tokens $\mathbf{t}^{clip}_i$ into the class-level embedding: $\mathbf{w}^{clip}_i=\theta(\mathbf{t}^{clip}_i)$.

To improve the discriminative of the class-level embedding, the prompt tuning methods of Context Optimization(CoOp) replaces the hand-crafted textual tokens with a set of learnable textual tokens $\mathbb{T}=\{\mathbf{t}_1, \mathbf{t}_2,...,\mathbf{t}_{M}\}$, where $M$ is the length of tokens.
Similar to CLIP, the corresponding class token $\mathbf{c}_i$ is concatenated with the learnable tokens $\mathbb{T}$ for generating the textual token $\mathbf{t}^{coop}_i=\{\mathbf{t}_1, \mathbf{t}_2,...,\mathbf{t}_{M},\mathbf{c}_i\}$.
Next, the textual embedding $\mathbf{w}^{coop}_{i}$ is obtained by fedding the textual tokens $\mathbf{t}^{coop}_i$ into Text Encoder $\theta$, \emph{i.e.}, $\mathbf{w}^{coop}_i=\theta(\mathbf{t}^{coop}_i)$.
Finally, the textual embeddings of all classes are defined as $\mathbf{W}^{coop}=\{\mathbf{w}^{coop}_i\}_{i=1}^{{N}_c}$.
CoOp infers the learnable textual tokens $\mathbb{T}$ by minimizing the contrastive loss between the image's embedding $\mathbf{x}$ and its class embedding $\mathbf{w}^{coop}_{y}$:
\begin{equation}
\mathcal{L}_{ce}=\frac{1}{N}\sum_{(\mathbf{x},y)\in \mathcal{D}_{s}}\frac{\exp(d(\mathbf{x},\mathbf{w}^{coop}_y)/\tau)}{\sum_{i=1}^{N_c}\exp(d(\mathbf{x},\mathbf{w}^{coop}_i)/\tau)},
\label{eq:coop}
\end{equation}
where $\mathcal{D}_{s}$ is the seen dataset, and $d(\cdot)$ is the cosine distance.
$\tau$ is a temperature factor defined in CLIP, and $N$ is the number of training images.

As the generated textual embedding has a good generalization ability for the novel classes, KgCoOp further adds an efficient consistency $\mathcal{L}_{kg}$ between the generated embedding $\mathbf{W}^{coop}$ and the general embedding $\mathbf{W}^{clip}$,
\begin{equation}
\mathcal{L}_{kg}=||\mathbf{W}^{clip}-\mathbf{W}^{coop}||_{2}^{2}.
\end{equation}

Therefore, a robust objective for prompt tuning is:
\begin{equation}
\mathcal{L} = \mathcal{L}_{ce}+\omega\mathcal{L}_{kg},
\label{eq:L}
\end{equation}
where $\omega$ is set as 8.0 same as in KgCoOp~\cite{YaoZX23}.

\subsection{Textual-based Class-aware Prompt tuning}
\begin{figure}
  \centering
   \includegraphics[width=0.8\linewidth]{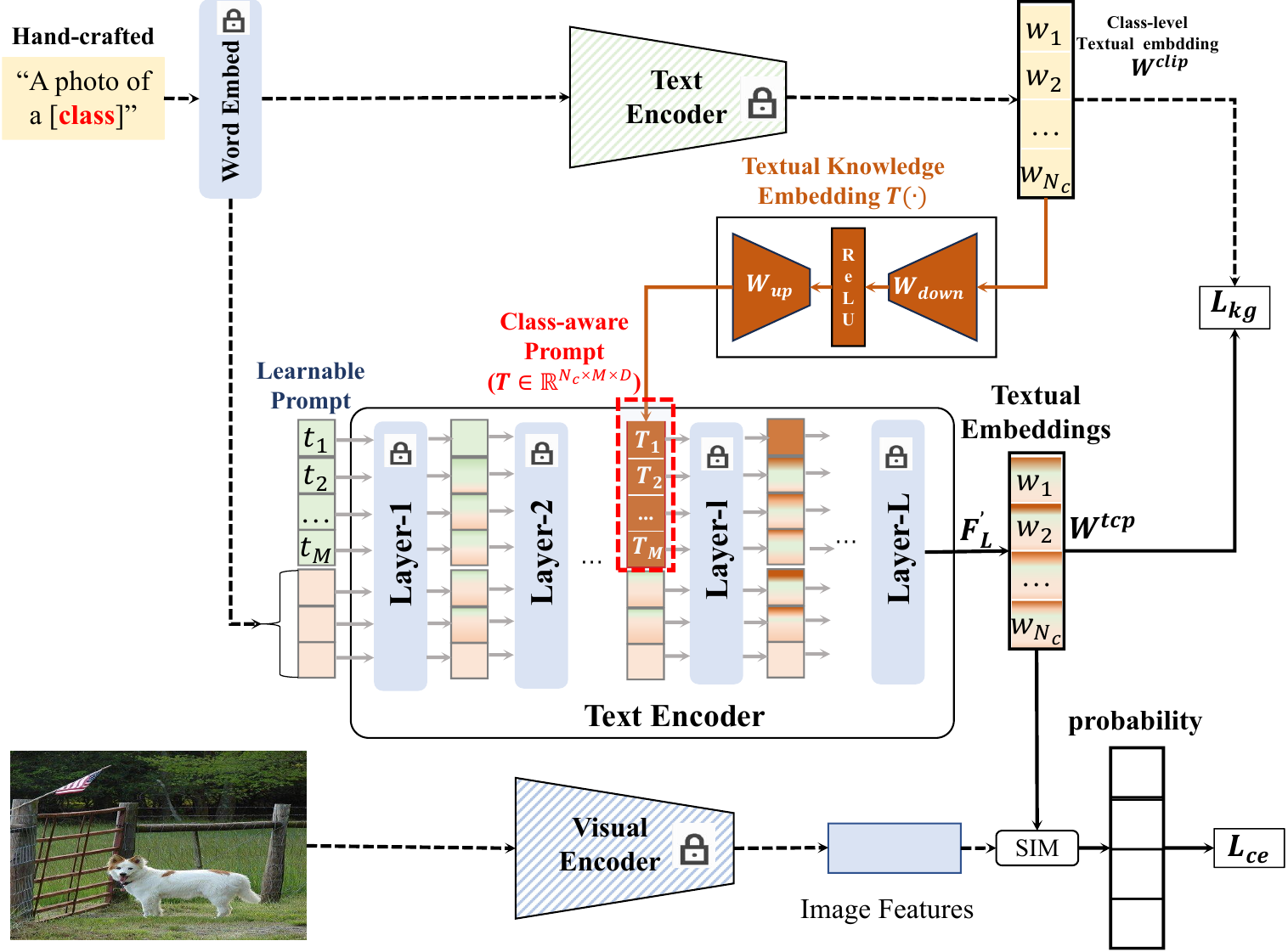}
   \caption{The framework of Textual-based Class-aware Prompt tuning.}
   \label{fig:TCP}
\vspace{-1.5em}
\end{figure}

Based on the pre-trained Text Encoder in CLIP, the textual prompt tuning aims to infer a set of the domain-shared or image-conditional textual tokens combined with the general class tokens for generating the specific class embeddings.
However, the textual classifier generated by those textual tokens performs a worse generalization on the unseen classes because they can not model the distribution of the testing classes.
Research has shown that utilizing the general textual knowledge extracted by frozen CLIP can create a discriminative prior knowledge of novel classes, enhancing the discriminative and generalization of the learnable prompt. 
Drawing on the benefit of the general textual knowledge of seen and unseen classes, we propose a Textual-based Class-aware Prompt tuning(TCP) to adapt the pretrained CLIP to the downstream task.
As shown in Figure~\ref{fig:TCP}, TCP employs a Textual Knowledge Embedding(TKE) to transfer the general class-level textual embedding into the class-aware prompt, which is then combined with the learnable textual tokens for generating the class-aware classifiers.
TKE is advantageous for unseen classes, as it generates class-specific prompts to obtain an unseen class-aware textual classifier with better discriminative abilities.
Moreover, explicitly incorporating seen class-aware prompts can enhance the discriminative power of seen classes. 

Given the general class-level textual embedding $\mathbf{W}^{clip}\in \mathbb{R}^{N_c\times D_t}$ with $N_c$ training classes, Textual Knowledge Embedding(TKE) $\mathcal{T}(\cdot)$ is proposed to project the class-level embedding $\mathbf{W}^{clip}$ into the class-aware prompt $\mathbf{T}$=$\mathcal{T}(\mathbf{W}^{clip})$.
As shown in Figure~\ref{fig:TCP}, TKE consists of two layers: down-project layer and up-project layer.
The down-project layer uses the weight $\mathbf{W}_{down}\in \mathbb{R}^{D_t\times D_{mid}}$ to project the textual embedding into the low-dimension feature with the dimension of $D_{mid}$.
Next, the weight $\mathbf{W}_{up}\in \mathbb{R}^{D_{mid}\times D'}$ of the up-project layer maps the low-dimension feature into the high-dimension feature with the dimension of $D'$.
Note that $D'$ is determined by the prompt's length $M$ and dimension $D$: $D'=M\times D$.
In summary, the general textual embedding $\mathbf{W}^{clip}\in \mathbb{R}^{N_c\times D_t}$ can be projected into the class-aware textual tokens $\mathbf{T}\in \mathbb{R}^{N_c\times D'}$, which is further reshaped into the shape of $\mathbf{T}\in\mathbb{R}^{N_c\times M \times D}$ for inserting into the middle layer of Text Encoder $\theta$. 

Assuming we insert the class-aware prompt $\mathbf{T}$ into $l$-th layer of Text Encoder $\theta$.
We will give a detailed analysis of the hyperparameter $l$ in the following.
Similar to CoOp, by combining the domain-shared learnable textual tokens $\mathbb{T}=[\mathbf{t}_1, \mathbf{t}_2,...,\mathbf{t}_{M}]$ and the pre-trained class tokens $\mathbf{C}$ of all classes, we can obtain the input textual tokens $\mathbf{F}_{0}=\{\mathbb{T},\mathbf{C}\}$ of the Text Encoder, where $\mathbf{C}=\{\mathbf{c}_i\}_{i}^{N_c}$, and $\mathbf{c}_i$ is the vectorized textual tokens of $i$-th class.
The textual token $\mathbf{F}_{0}$ is fed into the first $l$ layers of Text Encoder for obtaining the middle-level textual embedding $\mathbf{F}_{l}$. 
Formally, the textual token $\mathbf{F}_{i}(i\le l)$ of $i$-th layer is defined as:
\begin{equation}
\mathbf{F}_{i}=\theta_{i}(\mathbf{F}_{i-1}), i\in[1,l],
\end{equation}
where $\theta_{i}$ is the $i$-th layer of Text Encoder.


For the textual tokens $\mathbf{F}_{l}\in \mathbb{R}^{N_c\times N_t \times D}$ and the class-aware prompt tokens  $\mathbf{T}\in\mathbb{R}^{N_c\times M \times D}$, the first dimension is related to the number of classes.
Therefore, the learnable prompt tokens are always inserted into the second dimension of $\mathbf{F}_{l}$, the same as CoOp.
Formally, the class-aware prompt $\mathbf{T}$ is inserted into $\mathbf{F}_{l}$ to generate the class-aware enhanced tokens $\mathbf{F}'_{l}$,
\begin{equation}
\mathbf{F}'_{l}=[{\color{orange}\mathbf{T}_{1}},{\color{orange}\mathbf{T}_{2}},...,{\color{orange}\mathbf{T}_{M}},\mathbf{F}_{l,M+1},\mathbf{F}_{l,M+2},...,\mathbf{F}_{l,N_t}],
\end{equation}
where $\mathbf{T}_{i}$ denotes the $i$-th index of $\mathbf{T}$ in the second dimension, and $\mathbf{F}_{l,j}$ denote the correspond $j$-th index of $\mathbf{F}_{l}$ in the second dimension, \emph{i.e.,} $\mathbf{T}_{i}=\mathbf{T}[:,i,:]$, and  $\mathbf{F}_{l,j}=\mathbf{F}_{l}[:,j,:]$.

After that, the class-enhanced textual tokens $\mathbf{F}'_{l}$ are fedded into the rest layers for generating the class-aware textual embedding,
\begin{equation}
\mathbf{F}'_{i}=\theta_{i}(\mathbf{F}'_{i-1}), i\in[l+1,L].
\end{equation}

The output of the last layer $\mathbf{F}'_{L}$ is treated as the class embedding $\mathbf{W}^{tcp}$ used for optimization with contrastive loss and knowledge-guided consistency loss in Eq.~\eqref{eq:L}. 

\begin{table*}
\caption{Comparison on the base-to-new generalization setting with 16-shot samples.`\textcolor{orange}{tp}',`\textcolor{orange}{dtp}',`\textcolor{cyan}{vp}',and `\textcolor{cyan}{dvp}' denote the `\textcolor{orange}{textual prompt}', `\textcolor{orange}{deep textual prompt}',`\textcolor{cyan}{visual prompt}', and `\textcolor{cyan}{deep visual prompt}', respectively. PromptSRC are based on deep visual-textual prompt tuning(`\textcolor{cyan}{dvp}+\textcolor{orange}{dtp}'). `*' denote the performance obtained by our re-implementation.}
\scriptsize
\centering
\label{tab:base2new}
\begin{tabular}{lc|cccccccccccc|c}
\hline
\multirow{2}{*}{Datasets}   & \multirow{2}{*}{Sets} & \multirow{2}{*}{\begin{tabular}[c]{@{}c@{}}CoOp*\\ (\tiny IJCV22)\end{tabular}} & \multirow{2}{*}{\begin{tabular}[c]{@{}c@{}}CoCoOp\\ (\tiny CVPR22)\end{tabular}} & \multirow{2}{*}{\begin{tabular}[c]{@{}c@{}}DAPT*\\ (\tiny ICCV23)\end{tabular}} & \multirow{2}{*}{\begin{tabular}[c]{@{}c@{}}ProGrad*\\ (\tiny ICCV23)\end{tabular}} & \multirow{2}{*}{\begin{tabular}[c]{@{}c@{}}ProDA\\ (\tiny CVPR22)\end{tabular}} & \multirow{2}{*}{\begin{tabular}[c]{@{}c@{}}KgCoOp\\ (\tiny CVPR23)\end{tabular}} & \multirow{2}{*}{\begin{tabular}[c]{@{}c@{}}RPO\\ (\tiny ICCV23)\end{tabular}} & \multirow{2}{*}{\begin{tabular}[c]{@{}c@{}}PLOT*\\ (\tiny ICLR23)\end{tabular}} & \multirow{2}{*}{\begin{tabular}[c]{@{}c@{}}LFA\\ (\tiny ICCV23)\end{tabular}} & \multirow{2}{*}{\begin{tabular}[c]{@{}c@{}}MaPLe\\ (\tiny CVPR23)\end{tabular}} & \multirow{2}{*}{\begin{tabular}[c]{@{}c@{}}DePT\\ (\tiny Arxiv23)\end{tabular}} & \multirow{2}{*}{\begin{tabular}[c]{@{}c@{}}PromptSRC*\\ (\tiny ICCV23)\end{tabular}} & \textbf{TCP}   \\
                            &                       &                                                                          &                                                                            &                                                                          &                                                                             &                                                                           &                                                                            &                                                                         &                                                                          &                                                                         &                                                                           &                                                                           &                                                                               &       \\ \hline
                            &                       & \textcolor{orange}{tp}                                                                                                                                               & \textcolor{orange}{tp}                                                                         & \textcolor{orange}{tp}+\textcolor{cyan}{vp}                                                                    & \textcolor{orange}{tp}                                                                          & \textcolor{orange}{tp}                                                                        & \textcolor{orange}{tp}                                                                         & \textcolor{orange}{dtp}+\textcolor{cyan}{dvp}                                                                                                                                   & \textcolor{orange}{tp}+\textcolor{cyan}{vp}                                                                                                                                                                                                    & --                                                                      & \textcolor{orange}{dtp}+\textcolor{cyan}{dvp}                                                                                                                                   & \textcolor{orange}{tp}                                                                        & \textcolor{orange}{dtp}+\textcolor{cyan}{dvp}                                                                                                                                                                                                        & \textcolor{orange}{tp}    \\ \hline
\multirow{3}{*}{Average}    & Base                  & 82.38                                                                    & 80.47                                                                      & 83.18                                                                    & 82.48                                                                       & 81.56                                                                     & 80.73                                                                      & 81.13                                                                   & 83.98                                                                    & 83.62                                                                   & 82.28                                                                     & 83.62                                                                     & 84.12                                                                         & \textbf{84.13} \\
                            & New                   & 67.96                                                                    & 71.69                                                                      & 69.27                                                                    & 70.75                                                                       & 72.30                                                                     & 73.6                                                                       & 75.00                                                                   & 71.72                                                                    & 74.56                                                                   & 75.14                                                                     & 75.04                                                                     & 75.02                                                                         & \textbf{75.36} \\
                            & H                     & 74.48                                                                    & 75.83                                                                      & 75.59                                                                    & 76.16                                                                       & 76.65                                                                     & 77.0                                                                       & 77.78                                                                   & 77.37                                                                    & 78.83                                                                   & 78.55                                                                     & 79.10                                                                     & 79.31                                                                         & \textbf{79.51} \\ \hline
\multirow{3}{*}{ImageNet}   & Base                  & 76.46                                                                    & 75.98                                                                      & 76.83                                                                    & 77.02                                                                       & 75.40                                                                     & 75.83                                                                      & 76.60                                                                   & 77.30                                                                    & 76.89                                                                   & 76.66                                                                     & 77.03                                                                     & \textbf{77.75}                                                                         & 77.27 \\
                            & New                   & 66.31                                                                    & 70.43                                                                      & 69.27                                                                    & 66.66                                                                       & 70.23                                                                     & 69.96                                                                      & \textbf{71.57}                                                                   & 69.87                                                                    & 69.36                                                                   & 70.54                                                                     & 70.13                                                                     & 70.70                                                                         & 69.87 \\
                            & H                     & 71.02                                                                    & 73.10                                                                      & 72.85                                                                    & 71.46                                                                       & 72.72                                                                     & 72.78                                                                      & 74.00                                                                   & 73.40                                                                    & 72.93                                                                   & 73.47                                                                     & 73.42                                                                     & \textbf{74.06}                                                                         & 73.38 \\ \hline
\multirow{3}{*}{Caltech101} & Base                  & 97.80                                                                    & 97.96                                                                      & 97.83                                                                    & 98.02                                                                       & 98.27                                                                     & 97.72                                                                      & 97.97                                                                   & \textbf{98.53}                                                                    & 98.41                                                                   & 97.74                                                                     & 98.30                                                                     & 98.13                                                                         & 98.23 \\
                            & New                   & 93.27                                                                    & 93.81                                                                      & 93.07                                                                    & 93.89                                                                       & 93.23                                                                     & 94.39                                                                      & 94.37                                                                   & 92.80                                                                    & 93.93                                                                   & 94.36                                                                     & 94.60                                                                     & 93.90                                                                         & \textbf{94.67} \\
                            & H                     & 95.48                                                                    & 95.84                                                                      & 95.39                                                                    & 95.91                                                                       & 95.68                                                                     & 96.03                                                                      & 96.03                                                                   & 95.58                                                                    & 96.13                                                                   & 96.02                                                                     & 96.41                                                                     & 95.97                                                                         & \textbf{96.42} \\ \hline
\multirow{3}{*}{OxfordPets} & Base                  & 94.47                                                                    & 95.20                                                                      & 95.00                                                                    & 95.07                                                                       & 95.43                                                                     & 94.65                                                                      & 94.63                                                                   & 94.50                                                                    & 95.13                                                                   & 95.43                                                                     & 94.33                                                                     & \textbf{95.50}                                                                         & 94.67 \\
                            & New                   & 96.00                                                                    & 97.69                                                                      & 95.83                                                                    & 97.63                                                                       & \textbf{97.83}                                                                     & 97.76                                                                      & 97.50                                                                   & 96.83                                                                    & 96.23                                                                   & 97.76                                                                     & 97.23                                                                     & 97.40                                                                         & 97.20 \\
                            & H                     & 95.23                                                                    & 96.43                                                                      & 95.41                                                                    & 96.33                                                                       & \textbf{96.62}                                                                     & 96.18                                                                      & 96.05                                                                   & 95.65                                                                    & 95.68                                                                   & 96.58                                                                     & 95.76                                                                     & 96.44                                                                         & 95.92 \\ \hline
\multirow{3}{*}{Cars}       & Base                  & 75.67                                                                    & 70.49                                                                      & 75.80                                                                    & 77.68                                                                       & 74.70                                                                     & 71.76                                                                      & 73.87                                                                   & 79.07                                                                    & 76.32                                                                   & 72.94                                                                     & 79.13                                                                     & 78.40                                                                         & \textbf{80.80} \\
                            & New                   & 67.53                                                                    & 73.59                                                                      & 63.93                                                                    & 68.63                                                                       & 71.20                                                                     & 75.04                                                                      & \textbf{75.53}                                                                   & 74.80                                                                    & 74.88                                                                   & 74.00                                                                     & 75.47                                                                     & 74.73                                                                         & 74.13 \\
                            & H                     & 71.37                                                                    & 72.01                                                                      & 69.36                                                                    & 72.88                                                                       & 72.91                                                                     & 73.36                                                                      & 74.69                                                                   & 76.88                                                                    & 75.59                                                                   & 73.47                                                                     & 77.26                                                                     & 75.52                                                                         & \textbf{77.32} \\ \hline
\multirow{3}{*}{Flowers}    & Base                  & 97.27                                                                    & 94.87                                                                      & 96.97                                                                    & 95.54                                                                       & 97.70                                                                     & 95.00                                                                      & 94.13                                                                   & 97.93                                                                    & 97.34                                                                   & 95.92                                                                     & \textbf{98.00}                                                                     & 97.90                                                                         & 97.73 \\
                            & New                   & 67.13                                                                    & 71.75                                                                      & 60.90                                                                    & 71.87                                                                       & 68.68                                                                     & 74.73                                                                      & 76.67                                                                   & 73.53                                                                    & 75.44                                                                   & 72.46                                                                     & 76.37                                                                     & \textbf{76.77}                                                                         & 75.57 \\
                            & H                     & 79.44                                                                    & 81.71                                                                      & 74.81                                                                    & 82.03                                                                       & 80.66                                                                     & 83.65                                                                      & 84.50                                                                   & 83.99                                                                    & 85.00                                                                   & 82.56                                                                     & 85.84                                                                     & \textbf{86.06}                                                                         & 85.23 \\ \hline
\multirow{3}{*}{Food101}    & Base                  & 89.37                                                                    & \textbf{90.70}                                                                      & 90.37                                                                    & 90.37                                                                       & 90.30                                                                     & 90.5                                                                       & 90.33                                                                   & 89.80                                                                    & 90.52                                                                   & 90.71                                                                     & 90.50                                                                     & 90.63                                                                         & 90.57 \\
                            & New                   & 88.77                                                                    & 91.29                                                                      & 91.30                                                                    & 89.59                                                                       & 88.57                                                                     & 91.7                                                                       & 90.83                                                                   & 91.37                                                                    & 91.48                                                                   & \textbf{92.05}                                                                     & 91.60                                                                     & 91.50                                                                         & 91.37 \\
                            & H                     & 89.07                                                                    & 90.99                                                                      & 90.83                                                                    & 89.98                                                                       & 89.43                                                                     & 91.09                                                                      & 90.58                                                                   & 90.58                                                                    & 91.00                                                                   & \textbf{91.38}                                                                     & 91.05                                                                     & 91.06                                                                         & 90.97 \\ \hline
\multirow{3}{*}{Aircraft}   & Base                  & 39.67                                                                    & 33.41                                                                      & 39.97                                                                    & 40.54                                                                       & 36.90                                                                     & 36.21                                                                      & 37.33                                                                   & 42.13                                                                    & 41.48                                                                   & 37.44                                                                     & \textbf{43.20}                                                                     & 42.30                                                                         & 41.97 \\
                            & New                   & 31.23                                                                    & 23.71                                                                      & 29.80                                                                    & 27.57                                                                       & 34.13                                                                     & 33.55                                                                      & 34.20                                                                   & 33.73                                                                    & 32.29                                                                   & 35.61                                                                     & 34.83                                                                     & \textbf{36.97}                                                                         & 34.43 \\
                            & H                     & 34.95                                                                    & 27.74                                                                      & 34.14                                                                    & 32.82                                                                       & 35.46                                                                     & 34.83                                                                      & 35.70                                                                   & 37.46                                                                    & 36.31                                                                   & 36.50                                                                     & 38.57                                                                     & \textbf{39.46}                                                                         & 37.83 \\ \hline
\multirow{3}{*}{SUN397}     & Base                  & 80.85                                                                    & 79.74                                                                      & 80.97                                                                    & 81.26                                                                       & 78.67                                                                     & 80.29                                                                      & 80.60                                                                   & 82.20                                                                    & 82.13                                                                   & 80.82                                                                     & 82.33                                                                     & \textbf{82.83}                                                                         & 82.63 \\
                            & New                   & 68.34                                                                    & 76.86                                                                      & 76.97                                                                    & 74.17                                                                       & 76.93                                                                     & 76.53                                                                      & 77.80                                                                   & 73.63                                                                    & 77.20                                                                   & 78.70                                                                     & 77.80                                                                     & \textbf{79.00}                                                                         & 78.20 \\
                            & H                     & 74.07                                                                    & 78.27                                                                      & 78.92                                                                    & 77.55                                                                       & 77.79                                                                     & 78.36                                                                      & 79.18                                                                   & 77.68                                                                    & 79.59                                                                   & 79.75                                                                     & 80.00                                                                     & \textbf{80.87}                                                                         & 80.35 \\ \hline
\multirow{3}{*}{DTD}        & Base                  & 79.97                                                                    & 77.01                                                                      & 82.23                                                                    & 77.35                                                                       & 80.67                                                                     & 77.55                                                                      & 76.70                                                                   & 81.97                                                                    & 81.29                                                                   & 80.36                                                                     & 82.20                                                                     & 82.60                                                                         & \textbf{82.77} \\
                            & New                   & 48.60                                                                    & 56.00                                                                      & 54.23                                                                    & 52.35                                                                       & 56.48                                                                     & 54.99                                                                      & 62.13                                                                   & 43.80                                                                    & 60.63                                                                   & \textbf{59.18}                                                                     & 59.13                                                                     & 57.50                                                                         & 58.07 \\
                            & H                     & 60.46                                                                    & 64.85                                                                      & 65.36                                                                    & 62.45                                                                       & 66.44                                                                     & 64.35                                                                      & 68.61                                                                   & 57.09                                                                    & \textbf{69.46}                                                                   & 68.16                                                                     & 68.78                                                                     & 67.80                                                                         & 68.25 \\ \hline
\multirow{3}{*}{EuroSAT}    & Base                  & 90.10                                                                    & 87.49                                                                      & \textbf{94.73}                                                                    & 90.11                                                                       & 83.90                                                                     & 85.64                                                                      & 86.63                                                                   & 93.70                                                                    & 93.40                                                                   & 94.07                                                                     & 89.03                                                                     & 92.40                                                                         & 91.63 \\
                            & New                   & 53.00                                                                    & 60.04                                                                      & 50.33                                                                    & 60.89                                                                       & 66.00                                                                     & 64.34                                                                      & 68.97                                                                   & 62.67                                                                    & 71.24                                                                   & 73.23                                                                     & 71.07                                                                     & 68.43                                                                         & \textbf{74.73} \\
                            & H                     & 66.74                                                                    & 71.21                                                                      & 65.74                                                                    & 72.67                                                                       & 73.88                                                                     & 73.48                                                                      & 76.79                                                                   & 75.11                                                                    & 80.83                                                                   & 82.3                                                                     & 79.04                                                                     & 78.63                                                                         & \textbf{82.32} \\ \hline
\multirow{3}{*}{UCF101}     & Base                  & 84.53                                                                    & 82.33                                                                      & 84.30                                                                    & 84.33                                                                       & 85.23                                                                     & 82.89                                                                      & 83.67                                                                   & 86.60                                                                    & 86.97                                                                   & 83.00                                                                     & 85.80                                                                     & 86.93                                                                         & \textbf{87.13} \\
                            & New                   & 67.37                                                                    & 73.45                                                                      & 76.33                                                                    & 74.94                                                                       & 71.97                                                                     & 76.67                                                                      & 75.43                                                                   & 75.90                                                                    & 77.48                                                                   & 78.66                                                                     & 77.23                                                                     & 78.33                                                                         & \textbf{80.77} \\
                            & H                     & 74.98                                                                    & 77.67                                                                      & 80.12                                                                    & 79.35                                                                       & 78.04                                                                     & 79.65                                                                      & 79.34                                                                   & 80.90                                                                    & 81.95                                                                   & 80.77                                                                     & 81.29                                                                     & 82.41                                                                         & \textbf{83.83} \\ \hline
\end{tabular}
\vspace{-1.25em}
\end{table*}

\section{Experiments}
Similar to CoOp~\cite{ZhouYLL22}, we evaluate the effectiveness of TCP from the three types of tasks: 1) generalization from base-to-new classes within a dataset; 2)few-shot learning with $K$-shot labeled images; 3) cross-dataset generalization from the imagenet to other datasets.
\label{sec:exp}
\subsection{Experimental Setup}

\noindent\textbf{Dataset.}
We conduct the base-to-new generaliation, few-shot learning, and cross-dataset generalization on 11 benchmarks, \emph{i.e.,} ImageNet~\cite{DengDSLL009}, Caltech~\cite{Fei-FeiFP07}, OxfordPets~\cite{ParkhiVZJ12}, StanfordCars~\cite{Krause0DF13}, Flowers~\cite{NilsbackZ08}, Food101~\cite{BossardGG14}, FGVCAircraft~\cite{MajiRKBV13}, EuroSAT~\cite{HelberBDB19}, UCF101~\cite{abs-1212-0402}, DTD~\cite{CimpoiMKMV14}, and SUN397~\cite{XiaoHEOT10}.

\noindent\textbf{Training Details.}
Our implementation is based on CoOp's code~\cite{ZhouYLL22}.
All experiments are conducted based on the CLIP with the backbone of ViT-B/16~\cite{DosovitskiyB0WZ21}.
The prompt's length $M$ is set as 4 with random initialization of ``X X X X \{\}".
The final performance is averaged over three random seeds(1/2/3)for a fair comparison.
Adam optimizer is applied for optimization with the learning rate of 2e-3 and the batch size of 32, and the training epochs is 50.
All experiments are conducted on RTX 3090.

\noindent\textbf{Baselines.} Recently CoOp-based methods are used for comparison, \emph{e.g.,} CoOp~\cite{ZhouYLL22}, CoCoOp~\cite{ZhouYL022}, ProGrad~\cite{abs-2205-14865}, ProDA~\cite{proda}, KgCoOp~\cite{YaoZX23}, PromptSRC~\cite{DBLP:journals/corr/abs-2307-06948}, MaPLe~\cite{MAPLE23}, LFA~\cite{BlackBox}, DePT~\cite{DBLP:journals/corr/abs-2309-07439}, DAPT~\cite{DAPT23}, PLOT~\cite{0002YSLR023}, TaskRes~\cite{DBLP:conf/cvpr/0012LJ0W23}, RPO~\cite{RPO23}, VPT~\cite{abs-2210-07225}, and TIP-Adapter-F~\cite{DBLP:conf/eccv/ZhangZFGLDQL22}.

\subsection{Base-to-New Class Generalization}
To verify the generalization ability of the prompt tuning, the base-to-new class generalization splits each dataset into two disjoint subsets: \emph{Base} and \emph{New} classes, where \emph{Base} classes are used to infer the learnable prompt and TKE, and \emph{New} classes are used for evaluation.
Note that each \emph{New} class contains 16-shot samples.
We summarize the comparison between the proposed TCP and existing methods in Table~\ref{tab:base2new} from the following two aspects: Textual Prompt Tuning and Viusal-Textual Prompt Tuning.

\begin{table*}
\caption{Comparison of cross-dataset evaluation with existing methods.`\textcolor{orange}{tp}',`\textcolor{orange}{dtp}',`\textcolor{cyan}{vp}',and `\textcolor{cyan}{dvp}' denote the `\textcolor{orange}{textual prompt}', `\textcolor{orange}{deep textual prompt}',`\textcolor{cyan}{visual prompt}', and `\textcolor{cyan}{deep visual prompt}', respectively. Note that  DAPT and MaPLe are based on visual-textual prompt tuning(`\textcolor{cyan}{vp}+\textcolor{orange}{tp}').}
\label{tab:crossdataset}
\scriptsize
\centering
\begin{tabular}{l|cccccccccc|c}
\toprule
  Datasets            & CLIP  & CoOp  & ProGrad & KgCoOp & DePT           & VPT   & PLOT        & PromptSRC      & MaPLe   & DAPT           & \textbf{TCP}            \\
\midrule
              &       & \textcolor{orange}{tp}    & \textcolor{orange}{tp}      & \textcolor{orange}{tp}     & \textcolor{orange}{tp}             & \textcolor{orange}{tp}+\textcolor{cyan}{vp} & \textcolor{orange}{tp}+\textcolor{cyan}{vp}       & \textcolor{orange}{dtp}+\textcolor{cyan}{tvp}        & \textcolor{orange}{dtp}+\textcolor{cyan}{dvp} & \textcolor{orange}{tp}+\textcolor{cyan}{vp}          & \textcolor{orange}{tp}             \\
\midrule
ImageNet      & 66.70  & 71.51 & 72.24   & 70.66  & \textbf{72.77}          & 69.73 & 71.60        & 71.27          & 70.72   & 71.60  & 71.40           \\
\midrule
Caltech101    & 93.30  & 93.70 & 91.52   & 93.92  & \textbf{94.23} & 93.67 & 92.07       & 93.60          & 93.53   & 93.50           & 93.97          \\
OxfordPets    & 89.10  & 89.14 & 89.64   & 89.83  & 90.03          & 89.27 & 90.10        & 90.25          & 90.49   & 90.67          & \textbf{91.25} \\
StandfordCars & 65.70  & 64.51 & 62.39   & 65.41  & 65.57          & 65.5  & 65.70        & 65.70          & 65.57   & \textbf{65.93} & 64.69          \\
Flowers       & 70.70  & 68.71 & 67.87   & 70.01  & 70.57          & 70.2  & 69.23       & 70.25          & 72.20    & 71.70           & \textbf{71.21} \\
Food101       & 85.90  & 85.30 & 85.40   & 86.36  & 86.37          & 86.27 & 86.23       & 86.15          & 86.20   & 86.10           & \textbf{86.69} \\
FGVCAircraft  & 24.90  & 18.47 & 20.16   & 22.51  & 23.27          & 22.13 & \textbf{25.00} & 23.90          & 24.74   & 23.03          & 23.45          \\
SUN397        & 62.60  & 64.15 & 62.47   & 66.16  & 66.67          & 66.57 & 61.67       & 67.10          & 67.01   & 67.00             & \textbf{67.15} \\
DTD           & 44.30  & 41.92 & 39.42   & 46.35  & 45.97          & 46.93 & 38.60        & \textbf{46.87} & 46.49   & 44/00             & 44.35          \\
EuroSAT       & 48.30  & 46.39 & 43.46   & 46.04  & 43.53          & 47.43 & 47.83       & 45.50          & 48.06   & \textbf{52.47} & 51.45          \\
UCF101        & 67.60  & 66.55 & 64.29   & 68.50  & \textbf{69.30} & 67.20  & 67.00          & 68.75          & 68.69   & 68.73          & 68.73          \\
\midrule
Avg.          & 65.24 & 63.88 & 62.71   & 65.51  & 65.55          & 65.52 & 64.34      & 65.81          & 66.30   & \textbf{66.31} & 66.29         \\
\midrule 
\end{tabular}
\vspace{-1.25em}
\end{table*}

\emph{Textual Prompt Tuning:} Based on the visual feature extracted from the frozen visual encoder, the textual prompt tuning (\emph{e.g.,} CoOp, CoCoOp, ProGrad, ProDA, and KgCoOp) merely conducts the textual prompt tuning by inserting a set of learnable textual prompts into the Text Encoder for obtaining a discriminative textual classifier, which is same as our proposed TCP.
Among all textual prompt tuning methods, KgCoOp~\cite{YaoZX23} is a more substantial baseline, which is also the baseline of our TCP.
Compared to KgCoOp, TCP improves the metric term of \emph{Base}/\emph{New}/\emph{H} from 80.73\%/73.6\%/77.0\% to 84.35\%/75.28\%/79.56\%.
We can observe that a noticeable improvement of 3.62\% and 1.68\% are obtained in the \emph{Base} class and \emph{New} class, respectively.
The superior performance proves that TCP using the class-aware prompt to inject the class knowledge can generate the discriminative textual classifier for both seen and unseen classes.
DePT~\cite{DBLP:journals/corr/abs-2309-07439} is another work that also treats KgCoOp as the baseline and obtains the best \emph{H} of 79.10\% among all textual prompt tuning methods.
Compared to DePT, the proposed TCP obtains the performance of 84.35\%/75.28\%/79.56\% in the metric term of \emph{Base}/\emph{New}/\emph{H}, achieving the improvement of 0.51\%/0.32\%/0.41\%.

\emph{Visual-Textual Prompt Tuning:} Different from the textual prompt tuning merely adapts Text Encoder, DAPT~\cite{DAPT23}, PLOT~\cite{0002YSLR023}, RPO~\cite{RPO23}, MaPLe~\cite{MAPLE23}, and PromptSRC~\cite{DBLP:journals/corr/abs-2307-06948} all jointly performs prompt tuning on the text and visual encoders for both improving the discriminative of textual embedding and visual embeddings.
Unlike CoOp and TCP, which conduct the prompt tuning on one or two layers, those methods conduct the visual-textual prompt tunings on several layers of visual and text encoders.
Compared to them, the proposed TCP with shallow textual prompt tuning obtains a higher performance, \emph{e.g.,} 79.51\% \emph{vs} 79.31\% for PromptSRC in terms of \emph{H}.
Compared to deep visual-textual prompt tuning, the superior performance demonstrates that class-aware prompt tuning can generate discriminative and generalization textual-level classifiers.

\subsection{Cross-Dataset Generalization}
In the \emph{Base}-\emph{to}-\emph{New} generalization setting, \emph{New} classes always have a similar data distribution as the \emph{Base} classes.
To further verify the generalization of the proposed TCP, in the Cross-Dataset generalization, TCP is trained from the ImageNet and directly evaluated on the unrelated datasets, \emph{e.g.,} the rest ten datasets.
The comparison between the proposed TCP and existing methods is summarized in Table~\ref{tab:crossdataset}.
From Table~\ref{tab:crossdataset}, we can observe that the proposed TCP obtains the highest average performance among all textual prompt tuning methods(66.29\% \emph{vs} 65.55\% of DePT~\cite{DBLP:journals/corr/abs-2309-07439}), and obtains a comparable performance with the visual-textual prompt tuning methods (66.29\% \emph{vs} 66.31\% of DAPT~\cite{DAPT23}), demonstrating the effectiveness of TCP in learning the generalization knowledge.

\begin{table}
\caption{Effect of TCP on CoOp, KgCoOp, ProGrad, PromptSRC, and DAPT in the base-to-new generalization setting regarding final average performance.}
\centering
\tiny
\label{tab:lkg}
\begin{tabular}{l|ccc}
\toprule
Methods & \emph{Base} & \emph{New} & \emph{H} \\
\midrule
\midrule
CoOp~\cite{ZhouYLL22} & 82.63 & 67.99 & 74.6 \\
CoOp+TKE & 83.10$(\uparrow \textcolor{red}{0.47})$ & 70.17$(\uparrow \textcolor{red}{2.18})$ & 76.09 $(\uparrow \textcolor{red}{1.49})$ \\
\midrule
KgCoOp~\cite{YaoZX23} & 80.73 & 73.6 & 77.00 \\
KgCoOp +TKE & 84.13 $(\uparrow \textcolor{red}{3.40})$& 75.36$(\uparrow \textcolor{red}{1.76})$ & 79.51 $(\uparrow \textcolor{red}{2.51})$\\
\midrule
ProGrad~\cite{abs-2205-14865} & 82.48 & 70.75 & 76.16 \\
ProGrad+TKE& 82.61 $(\uparrow \textcolor{red}{0.13})$& 72.91$(\uparrow \textcolor{red}{2.16})$& 77.46 $(\uparrow \textcolor{red}{1.30})$\\
\midrule
PromptSRC~\cite{DBLP:journals/corr/abs-2307-06948} & 82.07 & 74.83 & 78.03 \\
PromptSRC+TKE& 83.74 $(\uparrow \textcolor{red}{1.67})$& 75.85$(\uparrow \textcolor{red}{1.02})$& 79.60 $(\uparrow \textcolor{red}{1.57})$\\
\midrule
DAPT~\cite{DAPT23} & 83.18 & 69.27 & 75.59 \\
DAPT+TKE& 84.15 $(\uparrow \textcolor{red}{0.97})$& 74.87$(\uparrow \textcolor{red}{5.60})$& 79.24 $(\uparrow \textcolor{red}{3.65})$\\
\bottomrule
\end{tabular}
\vspace{-1.25em}
\end{table}

\subsection{Few-shot Classification}
To verify that the proposed TCP can infer the class-aware knowledge, the few-shot classification is conducted on all 11 datasets using K-shot labeled source images and evaluated on the standard testing domain with the same class space as the training classes.
The 4-shot setting comparison between the proposed TCP and existing methods is summarized in Table~\ref{tab:fsl}.
We can observe that the proposed TCP achieves a higher average performance than existing methods, \emph{i.e.,} obtaining the average performance of 76.32\%.

\begin{table*}
\caption{Comparison of few-shot learning with 4-shot samples.}
\label{tab:fsl}
\scriptsize
\centering
\begin{tabular}{l|ccccccccccc|c}
\toprule
              & CLIP  & CoOp  & CoCoOp & ProGrad & KgCoOp & MaPLe & TIP-Adapter-F & DAPT  & PromptSRC & PLOT  & TaskRes & \textbf{TCP}  \\
\midrule
ImageNet      & 66.70  & 69.37 & 70.55  & 70.21   & 70.19  & 70.67 & 70.78         & 70.80  & \textbf{70.80}     & 70.40  & 62.87   & 70.48 \\
Caltech101    & 93.30  & 94.44 & 94.98  & 94.93   & 94.65  & 94.30  & 94.77         & 94.23 & 94.77     & \textbf{95.13} & 94.67   & 95.00    \\
OxfordPets    & 89.10  & 91.30 & \textbf{93.01}  & 93.21   & 93.20  & 92.05 & 92.26         & 92.17 & 93.23     & 92.55 & 92.00   & 91.90  \\
StandfordCars & 65.70  & 72.73 & 69.10  & 71.75   & 71.98  & 68.70  & 74.42         & 74.40  & 71.83     & 74.93 & 75.90   & \textbf{76.30}  \\
Flowers       & 70.70  & 91.14 & 82.56  & 89.98   & 90.69  & 80.80  & 92.98         & 92.37 & 91.31     & 92.93 & 91.50   & \textbf{94.40}  \\
Food101       & 85.90  & 82.58 & 86.64  & 85.77   & 86.59  & \textbf{86.90}  & 86.18         & 83.60  & 86.06     & 86.46 & 86.03   & 85.3  \\
FGVCAircraft  & 24.90  & 33.18 & 30.87  & 32.93   & 32.47  & 29.03 & 35.49         & 32.47 & 32.80     & 35.29 & 33.80   & \textbf{36.20}  \\
SUN397        & 62.60  & 70.13 & 70.50  & 71.17   & 71.79  & 71.47 & 70.65         & 72.20  & \textbf{72.80}     & 70.42 & 72.70   & 72.11 \\
DTD           & 44.30  & 58.57 & 54.79  & 57.72   & 58.31  & 54.73 & 61.70          & 61.37 & 60.64     & 62.43 & 59.57   & \textbf{63.97} \\
EuroSAT       & 48.30  & 68.62 & 63.83  & 70.84   & 71.06  & 54.87 & 78.27         & 72.73 & 75.02     & \textbf{80.70}  & 72.87   & 77.43 \\
UCF101        & 67.60  & 77.41 & 74.99  & 77.82   & 78.40  & 73.70 & 79.73         & 79.40  & 79.35     & 79.76 & 76.10   & \textbf{80.83} \\
\midrule
Avg.          & 65.37 & 73.59 & 71.98  & 74.21   & 74.48  & 70.66 & 76.11         & 75.07 & 75.33     & 76.45 & 74.36   & \textbf{76.72} \\
\bottomrule
\end{tabular}
\vspace{-1.25em}
\end{table*}

\subsection{Ablation Study}

\noindent\textbf{Effect of TKE's generalization.}
The critical component of TCP is the class-aware prompt generated by TKE, which embeds the prior class knowledge and can be easily integrated with existing CoOp-based methods.
As shown in Table~\ref{tab:lkg}, CoOp, KgCoOp, ProGrad, PromptSRC, and DAPT have all demonstrated improved performance when TKE is used. Specifically, improvements of 1.49\%, 2.51\%, 1.30\%, 1.57\%, and 3.65\% were achieved for CoOp, KgCoOp, ProGrad, PromptSRC~\footnote{PromptSRC is re-implemented with the shallow visual-textual prompt tuning.}, and DAPT, respectively. This superior performance highlights TCP's plug-and-play functionality, allowing easy integration with existing methods.

\noindent\textbf{Domain-shared prompt \emph{vs} the class-aware prompt.}
It is common for existing methods to infer the domain-shared prompt tokens. 
However, TCP differs significantly from them in using a class-aware prompt to capture class-aware knowledge in the Text Encoder. 
We thus compare the domain-shared and class-aware prompts, and summarize the related results in Table~\ref{tab:cap}. 
We can observe that simply using the class-aware prompt yields a fantastic performance of 79.36\% in terms of \emph{H}, higher than existing methods. 
Additionally, as shown in Table~\ref{tab:cap}, using the class-aware prompt results in a noticeable improvement over the domain-shared prompt. 
For instance, the \emph{Base}/\emph{New} performance is improved from 80.73\%/73.6\% to 84.05\%/75.18\%. 
The superior performance proves the reasonableness and effectiveness of injecting the class-aware knowledge into the prompt tuning. 
By further considering the domain-shared and class-aware prompts, the final performance is reached at 79.51\% 

\begin{table}
\caption{Domain-shared prompt \emph{vs} Class-aware prompt.}
\centering
\scriptsize
\label{tab:cap}
\begin{tabular}{ccc|ccc}
\toprule
Baseline & Domain-Shared & Class-Aware & Base & New & H \\
\midrule
\midrule
$\surd$ &$\surd$ & & 80.73 & 73.6 & 77 \\
$\surd$ &&$\surd$& 84.05 & 75.18 & 79.36 \\
$\surd$ &$\surd$&$\surd$ & \textbf{84.13} & \textbf{75.36} & \textbf{79.51} \\
\bottomrule
\end{tabular}
\vspace{-1.25em}
\end{table}

\begin{figure}
  \centering
   \includegraphics[width=0.8\linewidth]{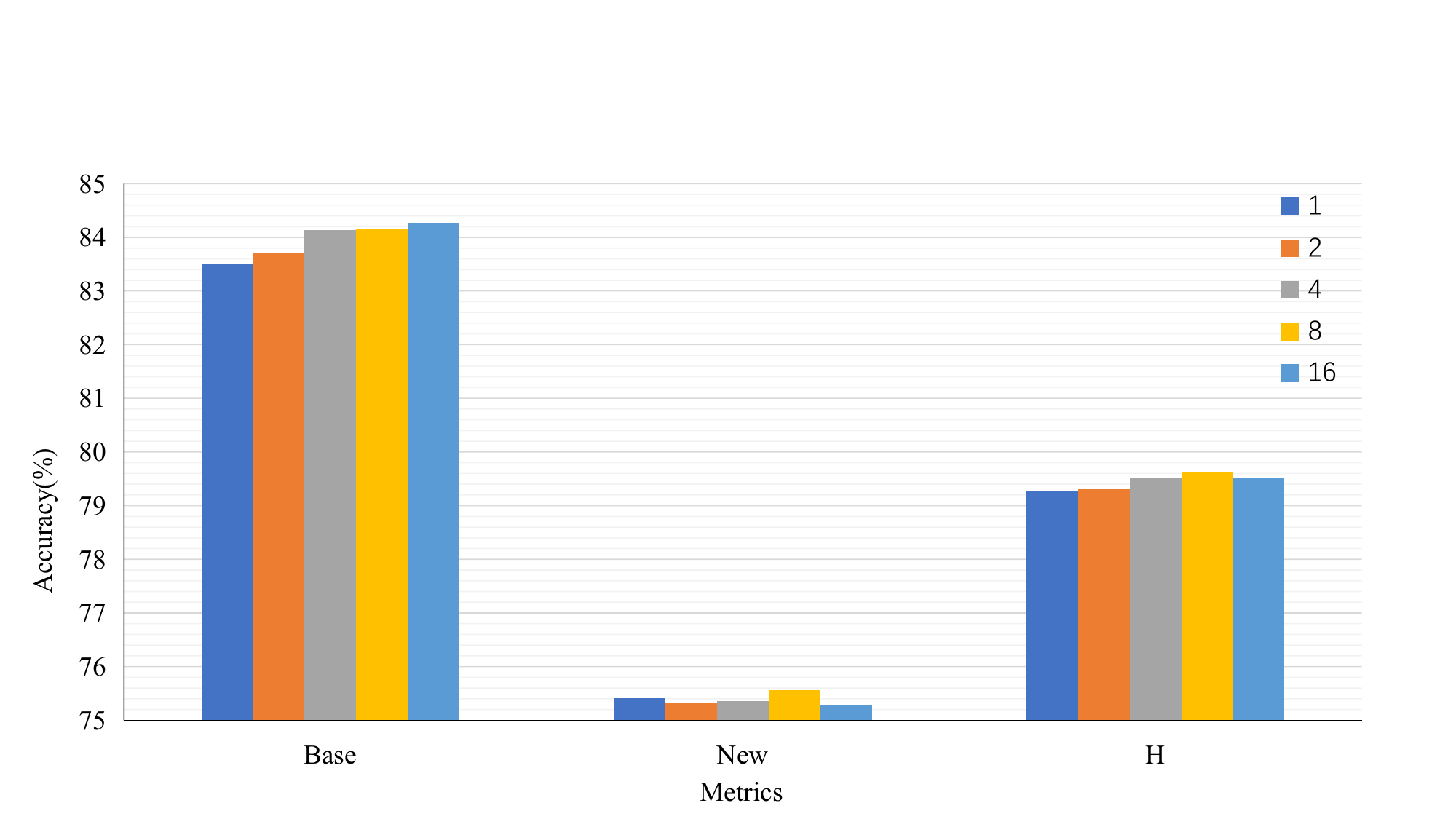}
   \caption{Effect of prompt length $M$ on the base-to-new generalization setting.}
   \label{fig:nctx}
\vspace{-1.25em}
\end{figure} 

\noindent\textbf{Effect of Prompt length $M$.}
In this study, we examine the impact of prompt length ($M$) on the \emph{Base}-\emph{to}-\emph{New} generalization setting. 
We compared prompt lengths of 1, 2, 4, 8, and 16, and the results are presented in Figure~\ref{fig:nctx}. 
The findings indicate that longer prompt lengths produce better performance, with $M=8$ producing the best results. 
However, the performance gap between the highest and lowest values of \emph{H}, which were 79.63\% and 79.26\%, respectively, is only 0.37\%. 
Therefore, the proposed TCP is relatively insensitive to prompt length ($M$).

\noindent\textbf{Effect of different templates.}
Unlike most CoOp-based methods using the handcrafted template ``a photo of a \{\}", the proposed TCP utilizes the random initialization template (``X X X X \{\}"). 
Therefore, we conducted a comparison among different templates.
Table~\ref{tab:init} shows that using the random template performs better than the handcrafted template. 
This is because TCP uses the class-ware prompt without considering the domain-share prompt, which has achieved better performance. 
For instance, in Table~\ref{tab:cap}, TCP with the class-aware prompt achieves comparable performance of 84.05\%/75.18\%/79.36\% in terms of \emph{Base}/\emph{New}/\emph{H} compared to existing methods. 
Moreover, the domain-shared prompt initialized as the random template provides complementary knowledge to the class-aware prompt.

\begin{table}
\caption{Comparison of different templates.}
\scriptsize
\centering
\label{tab:init}
\begin{tabular}{l|ccc}
\toprule
Templates & Base & New & H \\
\midrule
\midrule
 `X X X X {}'& \textbf{84.13} & \textbf{75.36} & \textbf{79.51} \\
`a photo of a {}'& 83.94 & 74.94 & 79.18 \\
 `this is a picture {}'& 84.00&74.66&79.06\\
\bottomrule
\end{tabular}
\vspace{-1.0em}
\end{table}

\begin{figure}
\begin{minipage}{0.485\linewidth}
\flushleft
\includegraphics[width=1.0\textwidth]{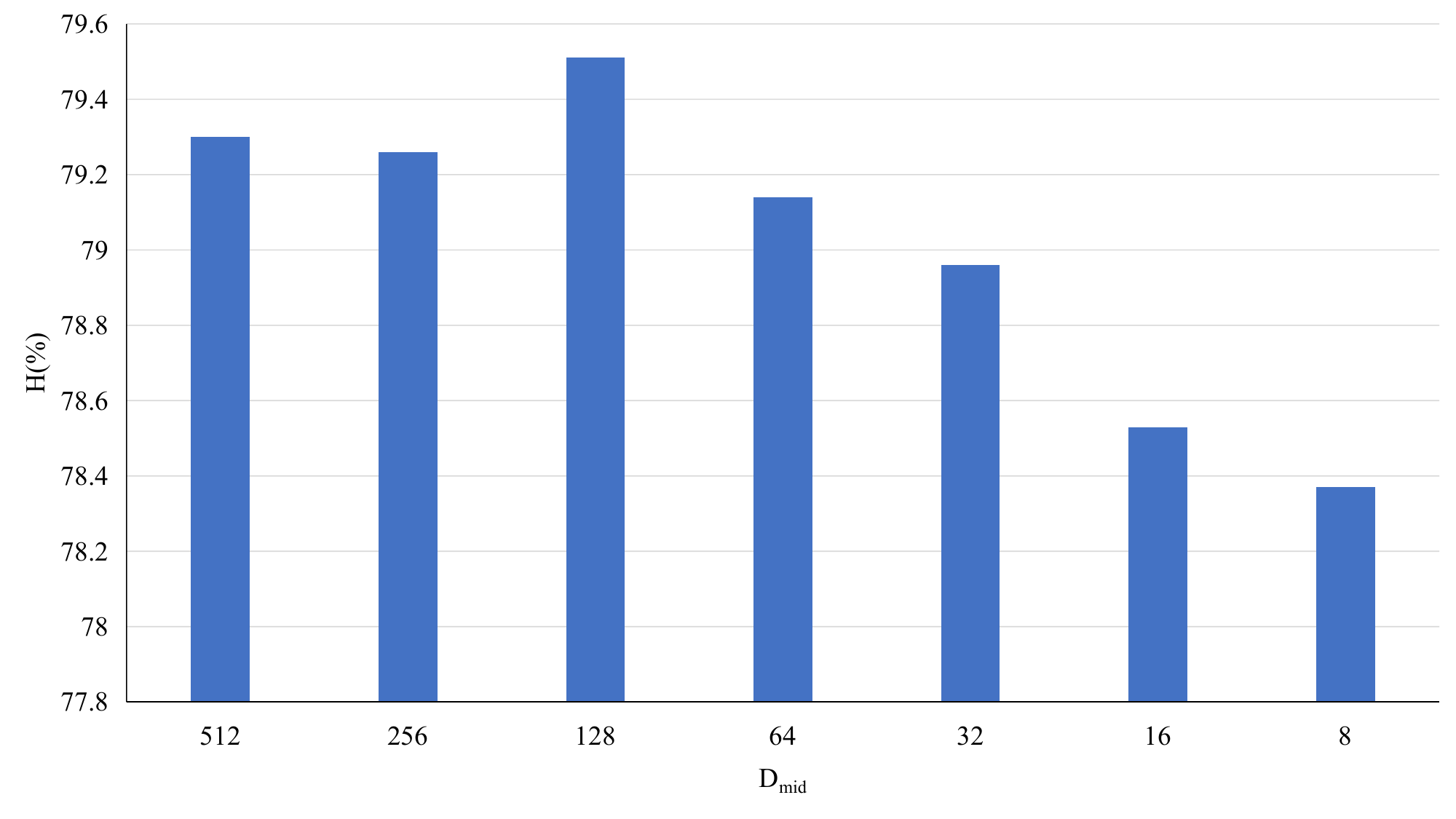}
\caption{Effect of dimension $D_{mid}$ in TKE.}
   \label{fig:D}
\end{minipage}
\begin{minipage}{0.485\linewidth}
	\flushright
   \includegraphics[width=1.0\textwidth]{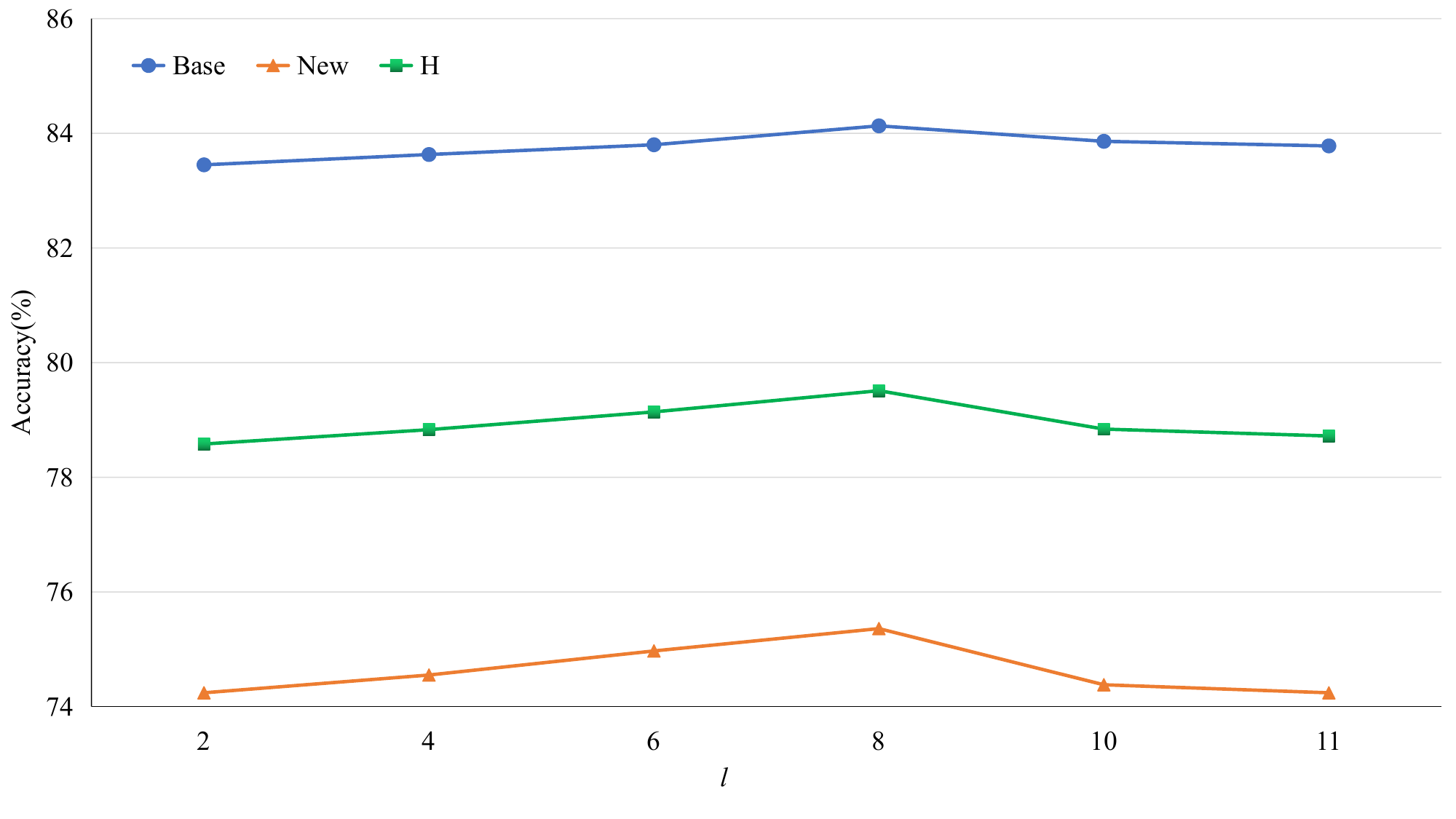}
   \caption{Effect of insert layer $l$.}
   \label{fig:layer}
\end{minipage}
\vspace{-1.0em}
\end{figure}

\noindent\textbf{Effect of $D_{mid}$ in TAK.}
We analyze the effect of the dimension $D_{mid}$ in TKE, and show the comparison in Figure~\ref{fig:D}.
Setting $D_{mid}$=128 obtains the best performance, and a smaller $D_{mid}$ would get a worse performance.

\noindent\textbf{Insert which layer?}
TCP inserts the class-aware textual prompts into a particular layer of the Text Encoder. 
The impact of inserting prompts into different layers was analyzed and shown in Figure~\ref{fig:layer}. 
The best performance was achieved when the class-aware prompt tuning was done on the 8th layer. 
This is because doing so on earlier layers ($l<8$) does not transfer class-aware knowledge to the textual embedding, and doing it on later layers can easily affect the textual embedding.

\noindent\textbf{Single-layer \emph{vs} Multiple-layers?}
The results of existing MaPLe and PromptSRC show that deep prompt learning applied on multiple layers can perform better than shallow prompt tuning with a single layer. 
Therefore, we compared the performance of shallow prompt tuning with deep prompt tuning. 
The results in Table~\ref{tab:multiple} indicate that the shallow class-aware prompt tuning with a single layer provides better performance. 
For instance, TCP-Shallow with 8-layer achieves the highest \emph{H} of 79.51\%, which surpasses all variants of TCP-deeper that have two, three, and four layers. 
Additionally, we observed that TCP-shallow obtains the best performance of 75.36\% for the \emph{New} class, and using more layers results in lower performance.

\begin{table}
\caption{\footnotesize Comparison with single-layer \emph{vs} multiple layers prompt tuning.}
\centering
\tiny
\label{tab:multiple}
\begin{tabular}{c|l|ccc}
\toprule
Modes &Layers & Base & New & H \\
\midrule
\midrule
TCP-Shallow&\{8\} 		&84.13	&\textbf{75.36}	&\textbf{79.51}\\
TCP-Deeper&\{4;8\} 		&84.02	&75.13	&79.33\\
TCP-Deeper&\{8;10\} 		&\textbf{84.24}	&75.29	&79.51\\
TCP-Deeper&\{4;8;10\} 	&84.11	&74.51	&79.02\\
TCP-Deeper&\{4;6;8;10\}	&84.17	&74.66	&79.13\\
\bottomrule
\end{tabular}
\vspace{-1.5em}
\end{table}

\begin{figure}
  \centering
   \includegraphics[width=0.85\linewidth]{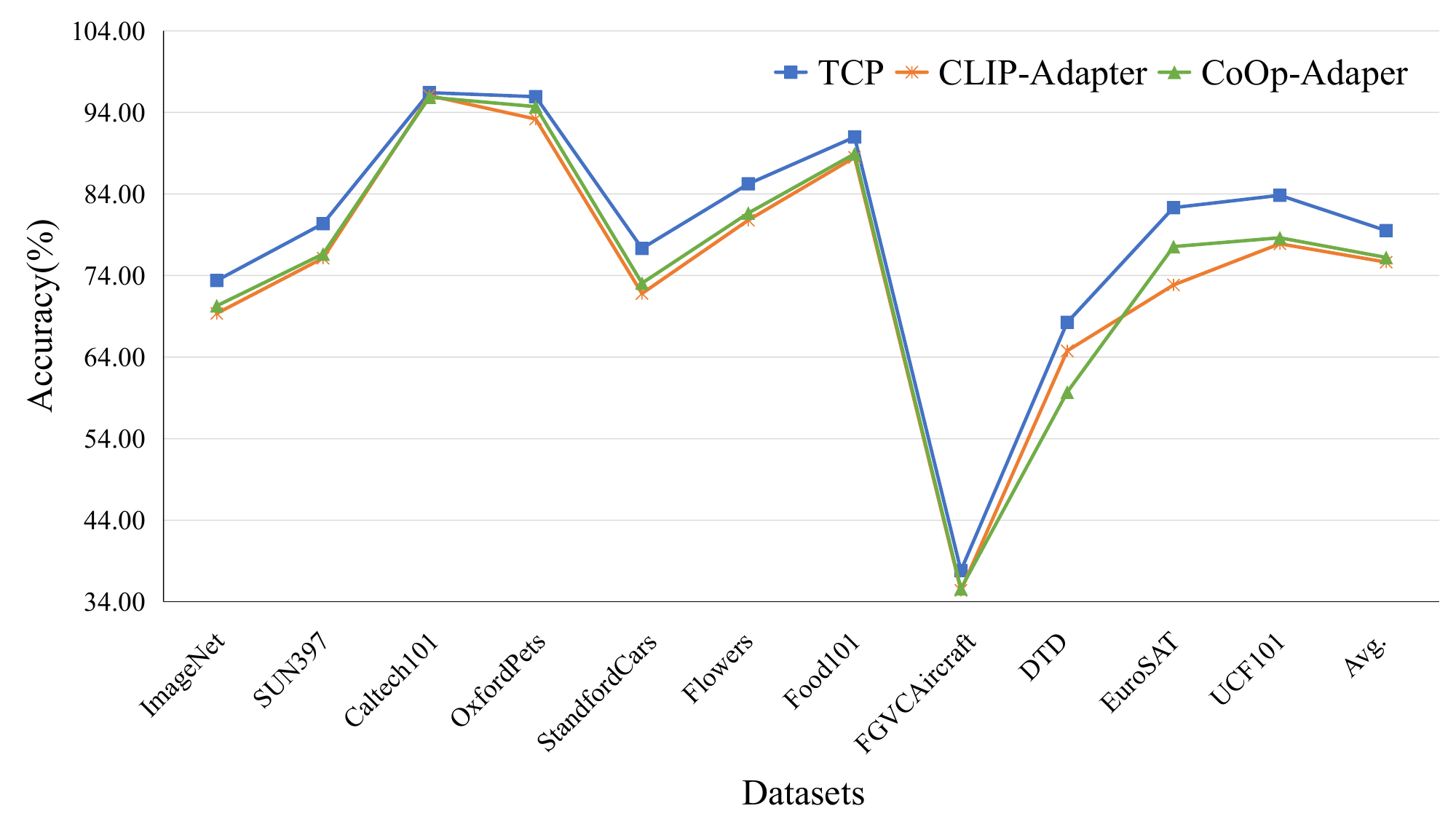}
   \caption{Comparison with the proposed TCP and Adapter-based methods.}
   \label{fig:adaper}
\vspace{-1.5em}
\end{figure}

\noindent\textbf{Compared to Adapter-based methods:}
In TCP, TKE acts as a special adapter, leading to the proposed TCP is similar to Adapter-based methods. 
We compared TCP with several Adapter-based methods such as CLIP-Adapter and CoOp-Adapter, which apply the adapter on the general textual-level class embeddings. 
However, TCP applies TKE to transfer the textual-level class embedding into the class-aware prompt combined with the class tokens in the middle layer of the Text Encoder. 
As shown in Figure~\ref{fig:adaper}, TCP with TKE performs better than the existing CLIP-Adapter and CoOp-Adapter on all 11 datasets.

\noindent\textbf{Training time analysis:}
We further analyze the training time of the proposed TCP. 
The training time was determined based on the average time to process one image from the ImageNet with a 16-shot setting. 
According to Table~\ref{tab:time}, the proposed TCP is an efficient method that performs better with less running time. 
This is because TCP adds an efficient TKE component for generating class-aware prompt upon CoOp and KgCoOp, resulting in slightly more time consumption than CoOp and KgCoOp.

\begin{table}
\caption{Training time comparison(ms/image).}
\centering
\tiny
\label{tab:time}
\begin{tabular}{c|ccccccc|c}
\toprule
	& CoOp & ProGrad & KgCoOp & PLOT & RPO & MaPLe & PromptSRC & TCP \\
\midrule
ms/image	& 6ms & 22ms & 6ms & 78.8ms  & 190ms & 90.7ms & 43.2ms & 6.4ms \\
H	& 74.48 & 76.16 & 77.00 & 77.37  & 77.78 & 78.55 & 79.31 & 79.51 \\
\bottomrule
\end{tabular}
\vspace{-1.5em}
\end{table}

\begin{figure}
  \centering
   \includegraphics[width=0.85\linewidth]{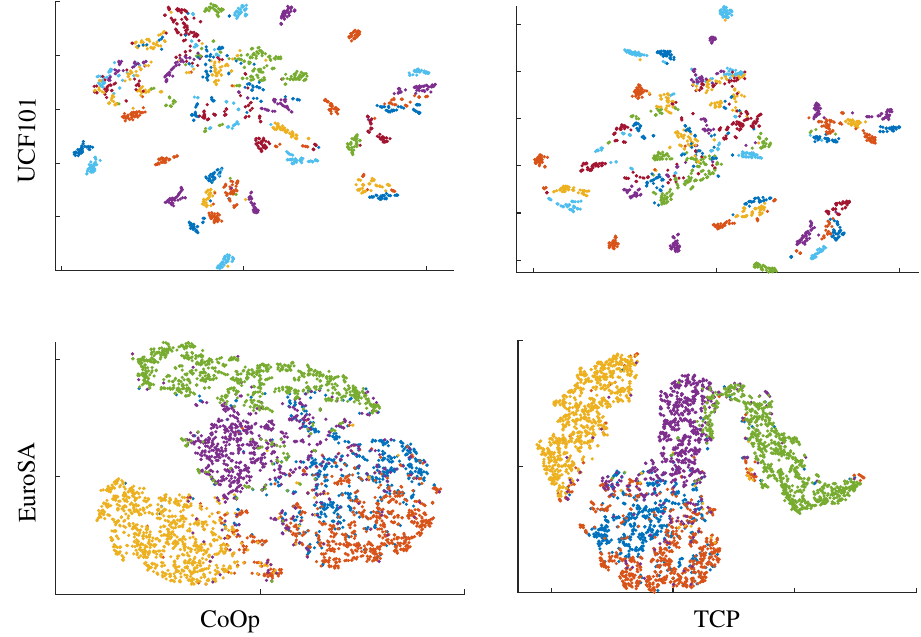}
   \caption{Visualization of the prediction probability obtained by CoOp and TCP.}
   \label{fig:tsne}
\vspace{-1.5em}
\end{figure}

\noindent\textbf{Visualization.} 
To prove that the proposed TCP can generate the discriminative textual classifier for prediction, we further visualize the prediction probability among all classes.
As shown in Figure~\ref{fig:tsne}, the proposed TCP has an obvious inter-class classes than existing methods.

\section{Conclusion}
To improve the generalization and discriminative abilities of the learnable prompt, we introduce a Textual-based Class-aware Prompt tuning method that takes advantage of the benefits of general class-level textual knowledge. To achieve this, we propose a Textual Knowledge Embedding (TKE) that transfers the class-level textual embedding into the class-aware prompt. This is combined with pre-trained class tokens to generate task-specific textual knowledge. Several benchmarks and tasks have shown that the class-aware prompt is effective for prompt tuning.

However, the class-aware prompt in TCP heavily relies on the discriminative ability of the general textual embedding. 
On the other hand, a weaker textual embedding will produce a weaker textual classifier. 
For example, TCP does not perform well on the FGVCAircraft dataset. Therefore, in the future, we plan to explore ways to use weaker textual knowledge to obtain a discriminative textual classifier.

\newpage
{\small
\bibliographystyle{ieee_fullname}
\bibliography{egbib}

\begin{thebibliography}{10}\itemsep=-1pt

\bibitem{ECO23}
Lorenzo Agnolucci, Alberto Baldrati, Francesco Todino, Federico Becattini,
  Marco Bertini, and Alberto Del~Bimbo.
\newblock Eco: Ensembling context optimization for vision-language models.
\newblock In {\em Proceedings of the IEEE/CVF International Conference on
  Computer Vision}, pages 2811--2815, 2023.

\bibitem{abs-2204-14198}
Jean-Baptiste Alayrac, Jeff Donahue, Pauline Luc, Antoine Miech, Iain Barr,
  Yana Hasson, Karel Lenc, Arthur Mensch, Katherine Millican, Malcolm Reynolds,
  et~al.
\newblock Flamingo: a visual language model for few-shot learning.
\newblock {\em Advances in Neural Information Processing Systems},
  35:23716--23736, 2022.

\bibitem{BossardGG14}
Lukas Bossard, Matthieu Guillaumin, and Luc~Van Gool.
\newblock Food-101 - mining discriminative components with random forests.
\newblock In David~J. Fleet, Tom{\'{a}}s Pajdla, Bernt Schiele, and Tinne
  Tuytelaars, editors, {\em Computer Vision - {ECCV} 2014 - 13th European
  Conference, Zurich, Switzerland, September 6-12, 2014, Proceedings, Part
  {VI}}, volume 8694 of {\em Lecture Notes in Computer Science}, pages
  446--461. Springer, 2014.

\bibitem{0002YSLR023}
Guangyi Chen, Weiran Yao, Xiangchen Song, Xinyue Li, Yongming Rao, and Kun
  Zhang.
\newblock {PLOT:} prompt learning with optimal transport for vision-language
  models.
\newblock In {\em The Eleventh International Conference on Learning
  Representations, {ICLR} 2023, Kigali, Rwanda, May 1-5, 2023}. OpenReview.net,
  2023.

\bibitem{DAPT23}
Eulrang Cho, Jooyeon Kim, and Hyunwoo~J Kim.
\newblock Distribution-aware prompt tuning for vision-language models.
\newblock In {\em Proceedings of the IEEE/CVF International Conference on
  Computer Vision}, pages 22004--22013, 2023.

\bibitem{CimpoiMKMV14}
Mircea Cimpoi, Subhransu Maji, Iasonas Kokkinos, Sammy Mohamed, and Andrea
  Vedaldi.
\newblock Describing textures in the wild.
\newblock In {\em 2014 {IEEE} Conference on Computer Vision and Pattern
  Recognition, {CVPR} 2014, Columbus, OH, USA, June 23-28, 2014}, pages
  3606--3613. {IEEE} Computer Society, 2014.

\bibitem{DengDSLL009}
Jia Deng, Wei Dong, Richard Socher, Li{-}Jia Li, Kai Li, and Li Fei{-}Fei.
\newblock Imagenet: {A} large-scale hierarchical image database.
\newblock In {\em 2009 {IEEE} Computer Society Conference on Computer Vision
  and Pattern Recognition {(CVPR} 2009), 20-25 June 2009, Miami, Florida,
  {USA}}, pages 248--255. {IEEE} Computer Society, 2009.

\bibitem{DosovitskiyB0WZ21}
Alexey Dosovitskiy, Lucas Beyer, Alexander Kolesnikov, Dirk Weissenborn,
  Xiaohua Zhai, Thomas Unterthiner, Mostafa Dehghani, Matthias Minderer, Georg
  Heigold, Sylvain Gelly, Jakob Uszkoreit, and Neil Houlsby.
\newblock An image is worth 16x16 words: Transformers for image recognition at
  scale.
\newblock In {\em 9th International Conference on Learning Representations,
  {ICLR} 2021, Virtual Event, Austria, May 3-7, 2021}. OpenReview.net, 2021.

\bibitem{Fei-FeiFP07}
Li Fei{-}Fei, Robert Fergus, and Pietro Perona.
\newblock Learning generative visual models from few training examples: An
  incremental bayesian approach tested on 101 object categories.
\newblock {\em Comput. Vis. Image Underst.}, 106(1):59--70, 2007.

\bibitem{abs-2210-09263}
Zhe Gan, Linjie Li, Chunyuan Li, Lijuan Wang, Zicheng Liu, Jianfeng Gao, et~al.
\newblock Vision-language pre-training: Basics, recent advances, and future
  trends.
\newblock {\em Foundations and Trends{\textregistered} in Computer Graphics and
  Vision}, 14(3--4):163--352, 2022.

\bibitem{abs-2110-04544}
Peng Gao, Shijie Geng, Renrui Zhang, Teli Ma, Rongyao Fang, Yongfeng Zhang,
  Hongsheng Li, and Yu Qiao.
\newblock Clip-adapter: Better vision-language models with feature adapters.
\newblock {\em CoRR}, abs/2110.04544, 2021.

\bibitem{DBLP:conf/iclr/HeZMBN22}
Junxian He, Chunting Zhou, Xuezhe Ma, Taylor Berg{-}Kirkpatrick, and Graham
  Neubig.
\newblock Towards a unified view of parameter-efficient transfer learning.
\newblock In {\em The Tenth International Conference on Learning
  Representations, {ICLR} 2022, Virtual Event, April 25-29, 2022}.
  OpenReview.net, 2022.

\bibitem{HeCXLDG22}
Kaiming He, Xinlei Chen, Saining Xie, Yanghao Li, Piotr Doll{\'{a}}r, and
  Ross~B. Girshick.
\newblock Masked autoencoders are scalable vision learners.
\newblock In {\em {IEEE/CVF} Conference on Computer Vision and Pattern
  Recognition, {CVPR} 2022, New Orleans, LA, USA, June 18-24, 2022}, pages
  15979--15988. {IEEE}, 2022.

\bibitem{HelberBDB19}
Patrick Helber, Benjamin Bischke, Andreas Dengel, and Damian Borth.
\newblock Eurosat: {A} novel dataset and deep learning benchmark for land use
  and land cover classification.
\newblock {\em {IEEE} J. Sel. Top. Appl. Earth Obs. Remote. Sens.},
  12(7):2217--2226, 2019.

\bibitem{HuSWALWWC22}
Edward~J. Hu, Yelong Shen, Phillip Wallis, Zeyuan Allen{-}Zhu, Yuanzhi Li,
  Shean Wang, Lu Wang, and Weizhu Chen.
\newblock Lora: Low-rank adaptation of large language models.
\newblock In {\em The Tenth International Conference on Learning
  Representations, {ICLR} 2022, Virtual Event, April 25-29, 2022}.
  OpenReview.net, 2022.

\bibitem{JiaYXCPPLSLD21}
Chao Jia, Yinfei Yang, Ye Xia, Yi{-}Ting Chen, Zarana Parekh, Hieu Pham,
  Quoc~V. Le, Yun{-}Hsuan Sung, Zhen Li, and Tom Duerig.
\newblock Scaling up visual and vision-language representation learning with
  noisy text supervision.
\newblock In Marina Meila and Tong Zhang, editors, {\em Proceedings of the 38th
  International Conference on Machine Learning, {ICML} 2021, 18-24 July 2021,
  Virtual Event}, volume 139 of {\em Proceedings of Machine Learning Research},
  pages 4904--4916. {PMLR}, 2021.

\bibitem{KAPT23}
Baoshuo Kan, Teng Wang, Wenpeng Lu, Xiantong Zhen, Weili Guan, and Feng Zheng.
\newblock Knowledge-aware prompt tuning for generalizable vision-language
  models.
\newblock In {\em Proceedings of the IEEE/CVF International Conference on
  Computer Vision}, pages 15670--15680, 2023.

\bibitem{MAPLE23}
Muhammad~Uzair Khattak, Hanoona Rasheed, Muhammad Maaz, Salman Khan, and
  Fahad~Shahbaz Khan.
\newblock Maple: Multi-modal prompt learning.
\newblock In {\em Proceedings of the IEEE/CVF Conference on Computer Vision and
  Pattern Recognition}, pages 19113--19122, 2023.

\bibitem{DBLP:journals/corr/abs-2307-06948}
Muhammad~Uzair Khattak, Syed~Talal Wasim, Muzammal Naseer, Salman Khan,
  Ming-Hsuan Yang, and Fahad~Shahbaz Khan.
\newblock Self-regulating prompts: Foundational model adaptation without
  forgetting.
\newblock In {\em Proceedings of the IEEE/CVF International Conference on
  Computer Vision}, pages 15190--15200, 2023.

\bibitem{KimSK21}
Wonjae Kim, Bokyung Son, and Ildoo Kim.
\newblock Vilt: Vision-and-language transformer without convolution or region
  supervision.
\newblock In Marina Meila and Tong Zhang, editors, {\em Proceedings of the 38th
  International Conference on Machine Learning, {ICML} 2021, 18-24 July 2021,
  Virtual Event}, volume 139 of {\em Proceedings of Machine Learning Research},
  pages 5583--5594. {PMLR}, 2021.

\bibitem{Krause0DF13}
Jonathan Krause, Michael Stark, Jia Deng, and Li Fei{-}Fei.
\newblock 3d object representations for fine-grained categorization.
\newblock In {\em 2013 {IEEE} International Conference on Computer Vision
  Workshops, {ICCV} Workshops 2013, Sydney, Australia, December 1-8, 2013},
  pages 554--561. {IEEE} Computer Society, 2013.

\bibitem{RPO23}
Dongjun Lee, Seokwon Song, Jihee Suh, Joonmyeong Choi, Sanghyeok Lee, and
  Hyunwoo~J Kim.
\newblock Read-only prompt optimization for vision-language few-shot learning.
\newblock In {\em Proceedings of the IEEE/CVF International Conference on
  Computer Vision}, pages 1401--1411, 2023.

\bibitem{DBLP:journals/corr/abs-2303-06571}
Juncheng Li, Minghe Gao, Longhui Wei, Siliang Tang, Wenqiao Zhang, Mengze Li,
  Wei Ji, Qi Tian, Tat-Seng Chua, and Yueting Zhuang.
\newblock Gradient-regulated meta-prompt learning for generalizable
  vision-language models.
\newblock {\em arXiv preprint arXiv:2303.06571}, 2023.

\bibitem{li2023blip}
Junnan Li, Dongxu Li, Silvio Savarese, and Steven Hoi.
\newblock Blip-2: Bootstrapping language-image pre-training with frozen image
  encoders and large language models.
\newblock {\em arXiv preprint arXiv:2301.12597}, 2023.

\bibitem{li2022blip}
Junnan Li, Dongxu Li, Caiming Xiong, and Steven Hoi.
\newblock Blip: Bootstrapping language-image pre-training for unified
  vision-language understanding and generation.
\newblock In {\em International Conference on Machine Learning}, pages
  12888--12900. PMLR, 2022.

\bibitem{LuBPL19}
Jiasen Lu, Dhruv Batra, Devi Parikh, and Stefan Lee.
\newblock Vilbert: Pretraining task-agnostic visiolinguistic representations
  for vision-and-language tasks.
\newblock In Hanna~M. Wallach, Hugo Larochelle, Alina Beygelzimer, Florence
  d'Alch{\'{e}}{-}Buc, Emily~B. Fox, and Roman Garnett, editors, {\em Advances
  in Neural Information Processing Systems 32: Annual Conference on Neural
  Information Processing Systems 2019, NeurIPS 2019, December 8-14, 2019,
  Vancouver, BC, Canada}, pages 13--23, 2019.

\bibitem{proda}
Yuning Lu, Jianzhuang Liu, Yonggang Zhang, Yajing Liu, and Xinmei Tian.
\newblock Prompt distribution learning.
\newblock In {\em {IEEE/CVF} Conference on Computer Vision and Pattern
  Recognition, {CVPR} 2022, New Orleans, LA, USA, June 18-24, 2022}, pages
  5196--5205. {IEEE}, 2022.

\bibitem{MajiRKBV13}
Subhransu Maji, Esa Rahtu, Juho Kannala, Matthew~B. Blaschko, and Andrea
  Vedaldi.
\newblock Fine-grained visual classification of aircraft.
\newblock {\em CoRR}, abs/1306.5151, 2013.

\bibitem{NilsbackZ08}
Maria{-}Elena Nilsback and Andrew Zisserman.
\newblock Automated flower classification over a large number of classes.
\newblock In {\em Sixth Indian Conference on Computer Vision, Graphics {\&}
  Image Processing, {ICVGIP} 2008, Bhubaneswar, India, 16-19 December 2008},
  pages 722--729. {IEEE} Computer Society, 2008.

\bibitem{DBLP:conf/cvpr/OquabBLS14}
Maxime Oquab, L{\'{e}}on Bottou, Ivan Laptev, and Josef Sivic.
\newblock Learning and transferring mid-level image representations using
  convolutional neural networks.
\newblock In {\em 2014 {IEEE} Conference on Computer Vision and Pattern
  Recognition, {CVPR} 2014, Columbus, OH, USA, June 23-28, 2014}, pages
  1717--1724. {IEEE} Computer Society, 2014.

\bibitem{BlackBox}
Yassine Ouali, Adrian Bulat, Brais Matinez, and Georgios Tzimiropoulos.
\newblock Black box few-shot adaptation for vision-language models.
\newblock In {\em Proceedings of the IEEE/CVF International Conference on
  Computer Vision}, pages 15534--15546, 2023.

\bibitem{ParkhiVZJ12}
Omkar~M. Parkhi, Andrea Vedaldi, Andrew Zisserman, and C.~V. Jawahar.
\newblock Cats and dogs.
\newblock In {\em 2012 {IEEE} Conference on Computer Vision and Pattern
  Recognition, Providence, RI, USA, June 16-21, 2012}, pages 3498--3505. {IEEE}
  Computer Society, 2012.

\bibitem{RadfordKHRGASAM21}
Alec Radford, Jong~Wook Kim, Chris Hallacy, Aditya Ramesh, Gabriel Goh,
  Sandhini Agarwal, Girish Sastry, Amanda Askell, Pamela Mishkin, Jack Clark,
  Gretchen Krueger, and Ilya Sutskever.
\newblock Learning transferable visual models from natural language
  supervision.
\newblock In Marina Meila and Tong Zhang, editors, {\em Proceedings of the 38th
  International Conference on Machine Learning, {ICML} 2021, 18-24 July 2021,
  Virtual Event}, volume 139 of {\em Proceedings of Machine Learning Research},
  pages 8748--8763. {PMLR}, 2021.

\bibitem{RaoZ0TZH0L22}
Yongming Rao, Wenliang Zhao, Guangyi Chen, Yansong Tang, Zheng Zhu, Guan Huang,
  Jie Zhou, and Jiwen Lu.
\newblock Denseclip: Language-guided dense prediction with context-aware
  prompting.
\newblock In {\em {IEEE/CVF} Conference on Computer Vision and Pattern
  Recognition, {CVPR} 2022, New Orleans, LA, USA, June 18-24, 2022}, pages
  18061--18070. {IEEE}, 2022.

\bibitem{schuhmann2022laion}
Christoph Schuhmann, Romain Beaumont, Richard Vencu, Cade Gordon, Ross
  Wightman, Mehdi Cherti, Theo Coombes, Aarush Katta, Clayton Mullis, Mitchell
  Wortsman, et~al.
\newblock Laion-5b: An open large-scale dataset for training next generation
  image-text models.
\newblock {\em Advances in Neural Information Processing Systems},
  35:25278--25294, 2022.

\bibitem{schuhmann2021laion}
Christoph Schuhmann, Richard Vencu, Romain Beaumont, Robert Kaczmarczyk,
  Clayton Mullis, Aarush Katta, Theo Coombes, Jenia Jitsev, and Aran
  Komatsuzaki.
\newblock Laion-400m: Open dataset of clip-filtered 400 million image-text
  pairs.
\newblock {\em arXiv preprint arXiv:2111.02114}, 2021.

\bibitem{DBLP:journals/corr/abs-2211-11720}
Sheng Shen, Shijia Yang, Tianjun Zhang, Bohan Zhai, Joseph~E. Gonzalez, Kurt
  Keutzer, and Trevor Darrell.
\newblock Multitask vision-language prompt tuning.
\newblock {\em CoRR}, abs/2211.11720, 2022.

\bibitem{singh2022flava}
Amanpreet Singh, Ronghang Hu, Vedanuj Goswami, Guillaume Couairon, Wojciech
  Galuba, Marcus Rohrbach, and Douwe Kiela.
\newblock Flava: A foundational language and vision alignment model.
\newblock In {\em Proceedings of the IEEE/CVF Conference on Computer Vision and
  Pattern Recognition}, pages 15638--15650, 2022.

\bibitem{abs-1212-0402}
Khurram Soomro, Amir~Roshan Zamir, and Mubarak Shah.
\newblock {UCF101:} {A} dataset of 101 human actions classes from videos in the
  wild.
\newblock {\em CoRR}, abs/1212.0402, 2012.

\bibitem{DBLP:journals/corr/abs-2303-15234}
Jingchen Sun, Jiayu Qin, Zihao Lin, and Changyou Chen.
\newblock Prompt tuning based adapter for vision-language model adaption.
\newblock {\em CoRR}, abs/2303.15234, 2023.

\bibitem{VaswaniSPUJGKP17}
Ashish Vaswani, Noam Shazeer, Niki Parmar, Jakob Uszkoreit, Llion Jones,
  Aidan~N. Gomez, Lukasz Kaiser, and Illia Polosukhin.
\newblock Attention is all you need.
\newblock In Isabelle Guyon, Ulrike von Luxburg, Samy Bengio, Hanna~M. Wallach,
  Rob Fergus, S.~V.~N. Vishwanathan, and Roman Garnett, editors, {\em Advances
  in Neural Information Processing Systems 30: Annual Conference on Neural
  Information Processing Systems 2017, December 4-9, 2017, Long Beach, CA,
  {USA}}, pages 5998--6008, 2017.

\bibitem{XiaoHEOT10}
Jianxiong Xiao, James Hays, Krista~A. Ehinger, Aude Oliva, and Antonio
  Torralba.
\newblock {SUN} database: Large-scale scene recognition from abbey to zoo.
\newblock In {\em The Twenty-Third {IEEE} Conference on Computer Vision and
  Pattern Recognition, {CVPR} 2010, San Francisco, CA, USA, 13-18 June 2010},
  pages 3485--3492. {IEEE} Computer Society, 2010.

\bibitem{YaoZX23}
Hantao Yao, Rui Zhang, and Changsheng Xu.
\newblock Visual-language prompt tuning with knowledge-guided context
  optimization.
\newblock In {\em Proceedings of the IEEE/CVF Conference on Computer Vision and
  Pattern Recognition}, pages 6757--6767, 2023.

\bibitem{DBLP:conf/cvpr/0012LJ0W23}
Tao Yu, Zhihe Lu, Xin Jin, Zhibo Chen, and Xinchao Wang.
\newblock Task residual for tuning vision-language models.
\newblock In {\em {IEEE/CVF} Conference on Computer Vision and Pattern
  Recognition, {CVPR} 2023, Vancouver, BC, Canada, June 17-24, 2023}, pages
  10899--10909. {IEEE}, 2023.

\bibitem{abs-2210-07225}
Yuhang Zang, Wei Li, Kaiyang Zhou, Chen Huang, and Chen~Change Loy.
\newblock Unified vision and language prompt learning.
\newblock {\em CoRR}, abs/2210.07225, 2022.

\bibitem{zhai2022scaling}
Xiaohua Zhai, Alexander Kolesnikov, Neil Houlsby, and Lucas Beyer.
\newblock Scaling vision transformers.
\newblock In {\em Proceedings of the IEEE/CVF Conference on Computer Vision and
  Pattern Recognition}, pages 12104--12113, 2022.

\bibitem{DBLP:journals/corr/abs-2309-07439}
Ji Zhang, Shihan Wu, Lianli Gao, Hengtao Shen, and Jingkuan Song.
\newblock Dept: Decoupled prompt tuning.
\newblock {\em arXiv preprint arXiv:2309.07439}, 2023.

\bibitem{DBLP:conf/eccv/ZhangZFGLDQL22}
Renrui Zhang, Wei Zhang, Rongyao Fang, Peng Gao, Kunchang Li, Jifeng Dai, Yu
  Qiao, and Hongsheng Li.
\newblock Tip-adapter: Training-free adaption of {CLIP} for few-shot
  classification.
\newblock In Shai Avidan, Gabriel~J. Brostow, Moustapha Ciss{\'{e}},
  Giovanni~Maria Farinella, and Tal Hassner, editors, {\em Computer Vision -
  {ECCV} 2022 - 17th European Conference, Tel Aviv, Israel, October 23-27,
  2022, Proceedings, Part {XXXV}}, volume 13695 of {\em Lecture Notes in
  Computer Science}, pages 493--510. Springer, 2022.

\bibitem{ZhouYL022}
Kaiyang Zhou, Jingkang Yang, Chen~Change Loy, and Ziwei Liu.
\newblock Conditional prompt learning for vision-language models.
\newblock In {\em {IEEE/CVF} Conference on Computer Vision and Pattern
  Recognition, {CVPR} 2022, New Orleans, LA, USA, June 18-24, 2022}, pages
  16795--16804. {IEEE}, 2022.

\bibitem{ZhouYLL22}
Kaiyang Zhou, Jingkang Yang, Chen~Change Loy, and Ziwei Liu.
\newblock Learning to prompt for vision-language models.
\newblock {\em Int. J. Comput. Vis.}, 130(9):2337--2348, 2022.

\bibitem{abs-2205-14865}
Beier Zhu, Yulei Niu, Yucheng Han, Yue Wu, and Hanwang Zhang.
\newblock Prompt-aligned gradient for prompt tuning.
\newblock In {\em Proceedings of the IEEE/CVF International Conference on
  Computer Vision}, pages 15659--15669, 2023.

\bibitem{PromptReg23}
Beier Zhu, Yulei Niu, Saeil Lee, Minhoe Hur, and Hanwang Zhang.
\newblock Debiased fine-tuning for vision-language models by prompt
  regularization.
\newblock In Brian Williams, Yiling Chen, and Jennifer Neville, editors, {\em
  Thirty-Seventh {AAAI} Conference on Artificial Intelligence, {AAAI} 2023,
  Thirty-Fifth Conference on Innovative Applications of Artificial
  Intelligence, {IAAI} 2023, Thirteenth Symposium on Educational Advances in
  Artificial Intelligence, {EAAI} 2023, Washington, DC, USA, February 7-14,
  2023}, pages 3834--3842. {AAAI} Press, 2023.

\end{thebibliography}
}

\newpage
\appendix

\section{Effect of Prompt Fusion}
As shown in Eq~\eqref{eq:prompt}(Eq.(5) in the paper), we insert the obtained class-aware prompt into the mid-level textual tokens by replacing $M$-th textual tokens as the class-aware prompt,
\begin{equation}
\label{eq:prompt}
\mathbf{F}'_{l}=[{\color{orange}\mathbf{T}_{1}},{\color{orange}\mathbf{T}_{2}},...,{\color{orange}\mathbf{T}_{M}},\mathbf{F}_{l,M+1},\mathbf{F}_{l,M+2},...,\mathbf{F}_{l,N_t}].
\end{equation}

Note that Eq~\eqref{eq:prompt} discards the mid-level textual tokens $\mathbf{\hat{F}}_{l}=[\mathbf{F}_{l,1},\mathbf{F}_{l,2},...,\mathbf{F}_{l,M}]$.
We thus reformuate Eq~\eqref{eq:prompt} by fusing the class-aware prompt  ${\color{orange}\mathbf{T}}$  and the discarded textual tokens $\mathbf{\hat{F}}_{l}$ with Eq.~\eqref{eq:prompt_fused},
\begin{equation}
\label{eq:prompt_fused}
\mathbf{F}'_{l}=[{\color{orange}\mathbf{T}'_{1}},{\color{orange}\mathbf{T}'_{2}},...,{\color{orange}\mathbf{T}'_{M}},\mathbf{F}_{l,M+1},\mathbf{F}_{l,M+2},...,\mathbf{F}_{l,N_t}],
\end{equation}
where ${\color{orange}\mathbf{T}'_{m}}$ is the $m$-th fused textual tokens, which is the combination of the class-aware prompt ${\color{orange}\mathbf{T}'_{m}}$ and the mid-level textual tokens $\mathbf{F}_{l,m}$,
\begin{equation}
\label{eq:fuse}
{\color{orange}\mathbf{T}'_{m}}=\omega{\color{orange}\mathbf{T}'_{m}}+(1-\omega)\mathbf{F}_{l,m},
\end{equation}
where $\omega$ is the weight.

\begin{figure}
  \centering
   \includegraphics[width=1.0\linewidth]{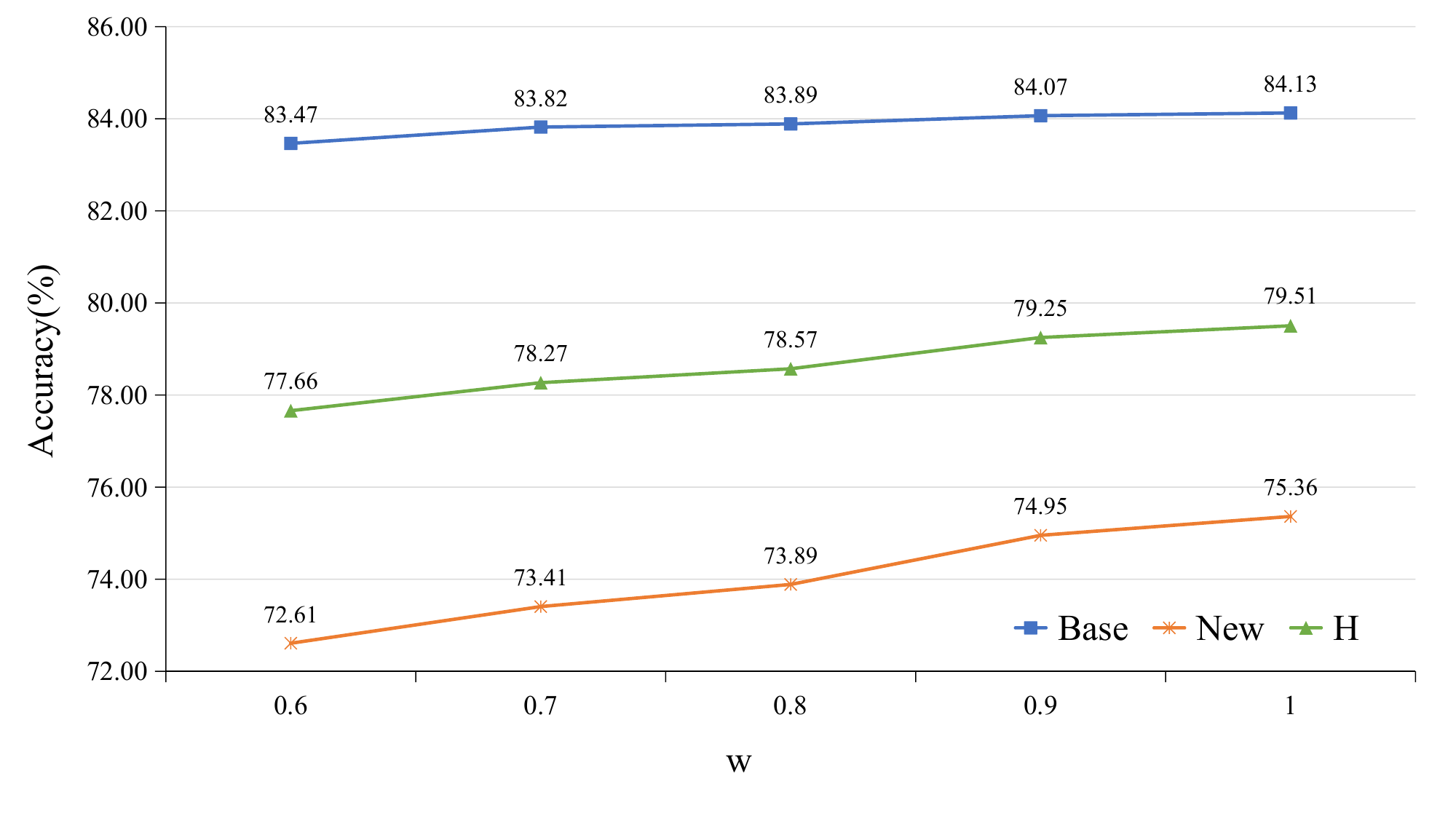}
   \caption{Effect of $\omega$ for prompt fusion.}
   \label{fig:w}
\end{figure}

We thus analyze the effect of $\omega$ for the TCP, and summarize the related results in Figure~\ref{fig:w}.
As shown in Figure~\ref{fig:w}, a higher weight $\omega$, a higher performance. 
Especially for the \emph{New} performance on the unseen classes, an obvious performance improvement is obtained using higher $\omega$.
The reason is that a higher $\omega$ in Eq.~\eqref{eq:fuse} means that a fewer mid-level textual tokens biased to the training domain is considerred for the testing domain.

\section{Class-aware Prompt \emph{vs} regularize $L_{kg}$}
The regularize term $L_{kg}$ constrains the \emph{output of TextEncoder}, while Class-aware Prompt explititly injects the class-related knowledge into the \emph{middle layer of TextEncoder}.
Moreover, class-aware prompt can explititly inject the class-level knowledge to increase the discrimiantive of textual-level classifier.
As shown in Table~\ref{tab:tcp}, combining the Class-aware Promptand $L_{kg}$ obtains a higher performance than merely using ones.

\begin{table}
\caption{Effect of class-aware prompt(CP)}
\label{tab:tcp}
\centering
\begin{tabular}{l|cc|ccc}
\toprule
&$L_{kg}$ & CP & Base & New & H\\
\midrule
CoOp & & & 82.38 & 67.96 & 74.48\\
KgCoOp &$\surd$ & & 80.73 & 73.6 & 77 \\
TCP* & & $\surd$& 82.99 & 73.07 & 77.72\\
TCP & $\surd$& $\surd$& 84.13 & 75.36 & 79.51 \\
\bottomrule
\end{tabular}
\end{table}

\begin{table*}
\small
\caption{Comparison of inference time and model complexity.}
\label{tab:time}
\centering
\begin{tabular}{l|ccccc}
\toprule
 &KgCoOp&CoOp&PromptSRC&MaPLe&TCP\\
\midrule
Total Parameters(M) &	124.325&124.325&124.369&127.887&124.654\\
\hline
Fixed Parameters with ViT-B/16($\mathcal{B}$)(M) &	&&124.323M&	&	\\
\hline
Learnable Parameters ($\mathcal{P}$)(M) &0.002&0.002&0.046&3.564&0.331\\
\hline
GFlops &1547.42&1547.42&1547.82&1547.84&1547.59\\
\hline
Time(ms/batch) &~124&~124&~124&~124&~124\\
\bottomrule
\end{tabular}
\end{table*}

\begin{table*}
\caption{Comparison of domain generalization.}
\label{tab:dg}
\centering
\begin{tabular}{l|ccccc|c}
\toprule
	& ImageNet & ImageNet-V2 & ImageNet-S & ImageNet-A & ImageNet-R & Avg.\\
\midrule
CoCoOp & 71.02 & 64.07 & 48.75 & 50.63& 76.18& 59.91 \\
ProGrad & 72.24 & 64.73 & 47.61 & 49.39 & 74.58 & 59.08 \\
KgCoOp & 71.2 & 64.1 & 48.97 & 50.69& 76.7 & 60.12 \\
MaPLe & 70.72 & 64.07 & 49.15 & 50.9 & 76.98 & 60.27\\
DAPT & 71.67 & 64.5 & 49.53 & 51.1 & 76.33 & 60.37 \\
TCP & 71.2 & 64.6 & 49.50 & 51.2 & 76.73 & \textbf{60.51} \\
\bottomrule
\end{tabular}
\end{table*}

\section{Comparison of training time and model complexity}
As the additional learable parameters($\mathcal{P}$) in prompt tuning is far smaller than the fixed parameters($\mathcal{B}$), the inferrece time is major controlled by the backbone.
Therefore, the methods with the same backbone(ViT-B/16) have the similar inference time(Tab.~\ref{tab:time}).

\begin{table*}
\centering
\caption{The detailed statistics of datasets used in our work.}
\small
\label{tab:dataset}
\begin{tabular}{l|c|ccc|c}
\toprule
Datasets & Classes & Training Size & Validation Size & Testing Size & Tasks \\
\midrule
\midrule
ImageNet~\cite{DengDSLL009}	&1,000	&1.28	M	&N/A 		&50,000	& General object recognition\\ 
Caltech~\cite{Fei-FeiFP07}		&100 		&4,128 	&1,649	&2,465 	& General object recognition\\
EuroSAT~\cite{HelberBDB19}			&10 		&13,500	&5,400	&8,100	& Satellite image recognition \\
SUN397~\cite{XiaoHEOT10}				&397		&15,880	&3,970	&19,850	&Scene recognition\\
DTD~\cite{CimpoiMKMV14}				&47		&2,820	&1,128	&1,692 	&Texture recognition\\
UCF101~\cite{abs-1212-0402}				&101		&7,639	&1,808	&3,783	&Action recognition\\
FGVCAircraft~\cite{MajiRKBV13}			&100 		&3,334	&3,333	&3,333	& Fine-grained aircraft recognition \\
OxfordPets~\cite{ParkhiVZJ12} 			&37		&2,944	&736		&3,669	& Fine-grained pets recognition\\
StanfordCars~\cite{Krause0DF13}			&196		&6,509	&1,635	&8,041	& Fine-grained car recognition\\
Flowers~\cite{NilsbackZ08}			&102		&4,093	&1,633	&2,463	&Fine-grained flowers recognition\\
Food101~\cite{BossardGG14}				&101		&50,500	&20,200	&30,300	& Fine-grained food recognition\\
\bottomrule
\end{tabular}
\end{table*}

\section{Domain Generalization}
Domain Generalization aims to evaluate the generalization by evaluating the model on the target dataset having the same class but different data distribution from the source domain.
Therefore, we conduct TCP on the few-shot ImageNets, and evaluate on the ImageNetV2, ImageNet-Sketch, ImageNet-A, and ImageNet-R.
The related results are summarized in Table~\ref{tab:dg}.

\section{Datasets}
Similar to existing CoOp-based methods, we conduct the evaluation on 11 datasets, \emph{i.e.,} ImageNet~\cite{DengDSLL009}, Caltech~\cite{Fei-FeiFP07}, OxfordPets~\cite{ParkhiVZJ12}, StanfordCars~\cite{Krause0DF13}, Flowers~\cite{NilsbackZ08}, Food101~\cite{BossardGG14}, FGVCAircraft~\cite{MajiRKBV13}, EuroSAT~\cite{HelberBDB19}, UCF101~\cite{abs-1212-0402}, DTD~\cite{CimpoiMKMV14}, and SUN397~\cite{XiaoHEOT10}.
As shown in Table~\ref{tab:dataset}, the type of datasets can be classified as: general object recognition, satellite image recognition, scene recognition, texture recognition, action recognition, and fine-grained object recognition.

\end{document}